\documentclass[11pt]{article}

\usepackage{microtype}
\usepackage{graphicx}
\usepackage{subcaption}
\usepackage{booktabs} % for professional tables
\usepackage{listings}
\usepackage{xcolor}

\usepackage{hyperref}
\usepackage{multirow}
\usepackage{enumitem}

\usepackage[margin=1in]{geometry}
\usepackage[T1]{fontenc}
\usepackage[utf8]{inputenc}
\usepackage{lmodern}
\usepackage{microtype}

\usepackage{amsmath,amssymb,amsthm}
\usepackage{graphicx}
\usepackage{booktabs}
\usepackage{multirow}
\usepackage{xcolor}
\usepackage{hyperref}
\usepackage[numbers]{natbib}

\usepackage{amsmath}
\usepackage{amssymb}
\usepackage{mathtools}
\usepackage{amsthm}
\usepackage{siunitx}

\usepackage[capitalize,noabbrev]{cleveref}

\lstdefinelanguage{YAML}{
  keywords={true,false,null,y,n},
  keywordstyle=\color{blue}\bfseries,
  basicstyle=\ttfamily\small,
  sensitive=false,
  commentstyle=\color{gray},
  stringstyle=\color{orange},
  morecomment=[l]{\#},
  breaklines=true, % IMPORTANT: automatic line breaking
  breakatwhitespace=true,
  columns=fullflexible,
}
\usepackage{authblk}

\theoremstyle{plain}

\theoremstyle{definition}

\theoremstyle{remark}

\usepackage[textsize=tiny]{todonotes}

\usepackage{makecell}

\newif\ifshowcomments
\newif\ifarxiv
\arxivfalse
\showcommentstrue % Set to true to show comments, set to false to hide
\ifshowcomments
    \newcommand{\jz}[1]{{\color{blue}[JZ: #1]}}
    \newcommand{\zhuozhao}[1]{{\color{red}[ZZ: #1]}}
    \newcommand{\ryt}[1]{{\color{purple}[RY: #1]}}
    \newcommand{\fn}[1]{{\color{green}[RY: #1]}}
\else
    \newcommand{\jz}[1]{}
    \newcommand{\zhuozhao}[1]{}
    \newcommand{\ryt}[1]{}
    \newcommand{\fn}[1]{}
   \ifarxiv
    \newcommand{\app}[1]{#1}
    \else
    \newcommand{\app}[1]{}
    \fi
\fi

\title{Contrastive Attribution in the Wild: An Interpretability Analysis of LLM Failures on Realistic Benchmarks}
\date{} % 去掉日期

\author{
  Rongyuan Tan$^{1}$\thanks{Work done during an internship at Microsoft.}\quad
  Jue Zhang$^{2}$\thanks{Corresponding authors.}\quad
  Zhuozhao Li$^{1\dagger}$\quad
  Qingwei Lin$^{2}$\quad
  Saravan Rajmohan$^{2}$\quad
  Dongmei Zhang$^{2}$
}

\date{
$^{1}$Southern University of Science and Technology, China\\
$^{2}$Microsoft\\
\texttt{juezhang@microsoft.com}, \texttt{lizz@sustech.edu.cn}
}

\begin{document}
\maketitle
\begin{abstract}
% https://chatgpt.com/share/6967606c-ffbc-8007-8bd1-01c579c09937
  %Interpretability tools are increasingly used to analyze failures of Large Language Models (LLMs), yet much of the existing work has focused on short prompts or toy settings. Consequently, it remains unclear how interpretability methods can be effectively applied to failure analysis in realistic settings. To address this gap, we examine existing interpretability approaches and identify contrastive, LRP-based attribution as a suitable tool for realistic failure analysis. We formulate failure analysis as \textit{contrastive attribution}, attributing the logit difference between an incorrect output token and a correct alternative to input tokens and internal model states, and introduce an efficient extension that enables construction of cross-layer hidden-state attribution graphs for long-context inputs. Using this framework, we conduct a systematic empirical study across multiple realistic benchmarks, comparing attribution patterns across datasets, model sizes and training checkpoints. Our results characterize how attribution-based interpretability behaves in realistic failure analysis and provide empirical grounding for its practical use.

  % https://chatgpt.com/share/69708a69-1670-8007-b430-23aba350409d
  Interpretability tools are increasingly used to analyze failures of Large Language Models (LLMs), yet prior work largely focuses on short prompts or toy settings, leaving their behavior on commonly used benchmarks underexplored. To address this gap, we study contrastive, LRP-based attribution as a practical tool for analyzing LLM failures in realistic settings. We formulate failure analysis as \textit{contrastive attribution}, attributing the logit difference between an incorrect output token and a correct alternative to input tokens and internal model states, and introduce an efficient extension that enables construction of cross-layer attribution graphs for long-context inputs. Using this framework, we conduct a systematic empirical study across benchmarks, comparing attribution patterns across datasets, model sizes, and training checkpoints. Our results show that this token-level contrastive attribution can yield informative signals in some failure cases, but is not universally applicable, highlighting both its utility and its limitations for realistic LLM failure analysis. Our code is available at: \url{https://aka.ms/Debug-XAI}.
\end{abstract}

\section{Introduction}

While the capabilities of Large Language Models (LLMs) have improved substantially in recent years, their deployment in real-world scenarios still reveals a wide range of failures~\cite{eval_survey_1}. 
Systematically debugging these failures is therefore crucial, both for understanding their underlying causes and for identifying actionable directions for further model improvement.

Existing analyses of LLM failures predominantly rely on \emph{behavioral analysis}, which examines model inputs, outputs, and responses to perturbations. Such analyses are effective at identifying failures, categorizing error types, and measuring sensitivity to prompt variations, but they primarily characterize \emph{what} fails rather than \emph{why} failures occur. A key limitation of this paradigm is its inherent ambiguity: similar behaviors can arise from distinct internal causes, such as underweighting relevant context tokens, overweighting irrelevant tokens, or biases embedded in internal representations. This motivates interpretability-based analysis~\cite{luo2024understandingutilizationsurveyexplainability}, which moves beyond behavior to reveal how models form preferences, attribute decisions, and propagate influence across layers.
%Disambiguating these causes is essential for effective model improvement and motivates interpretability-based analysis~\cite{luo2024understandingutilizationsurveyexplainability}, which goes beyond behavioral outcomes by exposing internal preference formation, attributing decisions to input tokens and hidden states, and tracing how influence propagates across model layers.

On the interpretability side, substantial progress has been made for transformer-based models~\cite{MI_survey}. However, applications to LLMs have largely been limited to toy settings, short inputs, or highly constrained prompt templates~\cite{DBLP:conf/iclr/WangVCSS23, rome, lindsey2025biology}. Interpretability analyses of realistic failures, such as those arising from standard benchmarks or real-world deployments, remain scarce. This limitation is partly attributable to the open-ended nature of such failures and their long input contexts, which can span thousands of tokens (e.g., in agentic tasks) and exceed the computational limits of many existing interpretability tools.\footnote{For example, the \textit{TransformerLens} package~\cite{nanda2022transformerlens} sets the default context length to 2048 tokens for the Qwen3 model series~\cite{yang2025qwen3technicalreport}.} Although recent work has begun to address scalability challenges in mechanistic interpretability~\cite{rosser2025streamscalingmechanisticinterpretability}, direct applications of interpretability methods to analyze realistic LLM failure cases remain limited, leaving their practical utility insufficiently understood.

Motivated by both the need for deeper failure debugging and the desire to establish the practical relevance of model interpretability, we investigate the use of interpretability methods for analyzing failures on \textit{standard benchmarks} (e.g., GAIA2~\cite{andrews2025arescalingagentenvironments}) as a first step toward more realistic settings. We focus on the central research question: \emph{Can interpretability analysis provide practical value for LLM failure case analysis?} In particular, we study scenarios that are difficult to resolve through behavioral analysis alone: (1) when a model produces an incorrect token, can interpretability reveal clues about the underlying decision process; (2) when a stronger model outperforms a weaker one, can interpretability expose evidence that the stronger model genuinely corrects the weaker model’s failure modes; and (3) across training checkpoints, can interpretability reflect the process by which models learn to make fewer mistakes.

To explore these questions, we formulate failure analysis as a problem of explaining why a model prefers an incorrect output token over a more correct alternative. We cast this as \textbf{Contrastive Attribution}~\cite{DBLP:conf/emnlp/YinN22}, attributing the logit difference between an incorrect token and a correct candidate to input tokens and internal model states. We build on \emph{AttnLRP}~\cite{DBLP:conf/icml/AchtibatHDJWLS24}, a state-of-the-art variant of the Layer-wise Relevance Propagation (LRP) framework~\cite{bach2015pixel}, and introduce an efficient extension that enables the construction of cross-layer hidden-state attribution graphs for long-context inputs. Using this framework, we conduct a systematic empirical study on hundreds of failure cases collected over multiple benchmarks, comparing attribution patterns across datasets, model sizes, and training checkpoints. Our results demonstrate that token-level contrastive attribution can yield informative signals for certain classes of failures, but is not universally effective, thereby clarifying both the promise and the limitations of interpretability methods for realistic LLM failure analysis.

Our contributions are summarized as follows:
\begin{itemize}[leftmargin=*]
    \item We formulate LLM failure analysis as a token-level \emph{contrastive attribution} problem, enabling the application of model interpretability methods to analyze failures in a more realistic setting than prior work.
    \item We introduce an efficient extension of LRP-based attribution that enables scalable construction of cross-layer hidden-state attribution graphs for long-context inputs.
    \item We conduct a systematic interpretability-based failure attribution analysis across common benchmarks, characterizing attribution patterns across datasets, model sizes, and training checkpoints in realistic failure scenarios.
\end{itemize}

\section{Related Work}

\paragraph{Analyzing LLM Failures.}
LLM failures are often encountered and studied after benchmark evaluation. Such analyses have revealed systematic weaknesses in areas including natural language inference, semantic and abstract reasoning, robustness, and bias~\cite{eval_survey_1}. However, these studies largely treat LLMs as black boxes, relying on output-level metrics without examining the internal mechanisms that lead to failures.

Another prominent line of research studies LLM failures under the umbrella of \emph{hallucination}.
%, broadly defined as non-factual, unsupported, or unfaithful generation. 
Empirical work in this area characterizes hallucination behaviors
%, for example by distinguishing intrinsic and extrinsic hallucinations
~\cite{maynez-etal-2020-faithfulness} as covered by recent surveys~\cite{hallu_survey_1,hallu_survey_2,hallu_survey_3}. To facilitate controlled analysis, several hallucination-focused benchmarks have been proposed~\cite{trufulQA,HaluEval,HalluLens}. While these datasets are valuable for isolating hallucination phenomena, the resulting failures are tightly coupled to dataset design and differ from errors that arise organically in standard benchmark evaluations.

Interpretability methods have also been applied to study hallucination-related behaviors, but predominantly in synthetic or simplified settings. Prior work~\cite{rome,dai-etal-2022-knowledge,geva-etal-2021-transformer,meng2023massediting,NEURIPS2024_d6df31b1,lindsey2025biology} often relies on short prompts or highly structured formats, such as subject--relation--object templates~\cite{rome}. These settings abstract away much of the complexity present in realistic, open-ended benchmark failures.

Research on LLM reasoning failures has also gained renewed attention with the emergence of large reasoning models~\cite{li202512surveyreasoning}. Recent studies categorize reasoning errors~\cite{song2025a}, monitor reasoning processes~\cite{baker2025monitoring}, steer or interpret reasoning via sparse features~\cite{galichin2025icoveredbaseshere}, and debug reasoning failures~\cite{zhang-etal-2025-reasoning}. However, most of these works focus primarily on reasoning tokens, rather than connecting failures to broader benchmark-defined errors.

Lastly, a growing body of research work examines failures in LLM-powered agent systems, including failure taxonomies~\cite{MAST}, failure localization~\cite{whoandwhen}, and agent debugging methods~\cite{ma2025doverinterventiondrivenautodebugging}. These studies primarily operate at the system or agent level and do not directly investigate the underlying causes of failure by incorporating the base LLMs.

\paragraph{Input Attribution.}
This work adopts LRP to attribute the final token logit difference to input tokens and internal states. More broadly, input attribution methods aim to localize model behavior by estimating the contribution of individual input tokens to model predictions~\cite{MI_survey}. A prominent class of approaches is gradient-based attribution, including gradient norms~\cite{DBLP:journals/corr/SimonyanVZ13}, $gradient \times input$~\cite{denil2015extractionsalientsentenceslabelled}, integrated gradients~\cite{DBLP:conf/icml/SundararajanTY17}, and SmoothGrad~\cite{DBLP:journals/corr/SmilkovTKVW17}. Another widely studied family comprises perturbation-based methods, which estimate input importance by adding noise or ablating input elements and measuring the resulting change in model predictions~\cite{DBLP:conf/nips/LundbergL17, DBLP:conf/iccv/FongV17, DBLP:journals/jmlr/CovertLL21}.

Beyond these, attention-based and context-mixing methods aim to decompose the final-layer token representations used for prediction into layer-wise contributions by explicitly tracking how attention mixes information across tokens~\cite{DBLP:conf/emnlp/KobayashiKYI21, DBLP:conf/acl/FerrandoGTC23}. Within the LRP framework, several variants beyond AttnLRP have also been proposed~\cite{DBLP:conf/icml/AliSEMMW22}. Among existing approaches, it was shown that AttnLRP achieved state-of-the-art faithfulness in input token attribution for transformer models~\cite{DBLP:conf/icml/AchtibatHDJWLS24}. We therefore adopt AttnLRP as the interpretability method in this work. 

Recent studies have also focused on improving the computational efficiency of attribution methods, which often incur substantial overhead~\cite{DBLP:conf/iclr/LiuKR25, arras2025closelookdecompositionbasedxaimethods}. Our efficient extension of AttnLRP builds on the backpropagation-based method in~\cite{arras2025closelookdecompositionbasedxaimethods}.

\paragraph{Causal Patching, Circuit Discovery, and Attribution Graph.}

Beyond input-level attribution, previous work has also investigated attribution to internal model states and the identification of critical computation paths (\emph{circuits}) that give rise to a target output. A prominent class of methods relies on causal patching, including activation patching~\cite{heimersheim2024useinterpretactivationpatching,DBLP:conf/iclr/ZhangN24}, attribution patching~\cite{nanda2023attribution, DBLP:journals/corr/abs-2403-00745}, and path patching~\cite{DBLP:conf/iclr/WangVCSS23, goldowskydill2023localizingmodelbehaviorpath}. These approaches often introduce paired \emph{clean} and \emph{corrupted} prompts and identify critical model components by selectively patching activations from one run into the other. While such contrastive setups provide strong causal control, constructing suitable clean--corrupted prompt pairs is often nontrivial, especially when analyzing real-world failures.

Another line of work trains sparse autoencoders (e.g., transcoders) to obtain locally linearized representations of model activations, enabling circuit discovery, attribution graph construction, and mechanistic analysis~\cite{DBLP:conf/nips/DunefskyCN24, ameisen2025circuit, zhao2025verifyingchainofthoughtreasoningcomputational, dai2025graphghosttracingstructureslarge}. Although these methods can reveal fine-grained, neuron-level structure, they are computationally intensive, typically restricted to short prompts, and require substantial human expertise for effective analysis. Moreover, the need to train a separate transcoder for each model limits their scalability across different LLMs.

%\jz{potentially to add a footnote that while there does exist transcoders for 0.6B, we cannot establish the attribution graph for the failures cases in our study due to long context.}

Attribution graphs can also be constructed using decomposition-based methods such as LRP and attention-based approaches~\cite{DBLP:conf/emnlp/FerrandoV24, rosser2025streamscalingmechanisticinterpretability}. These techniques are generally more computationally efficient while maintaining reasonable faithfulness in their attributions. Given the long-context requirements inherent in realistic failure analysis, we therefore develop attribution graphs based on AttnLRP in this work.

\section{Methodology}

%In this section, we first present a formal problem formulation for contrastive attribution. We then introduce an efficient extension of AttnLRP for constructing attribution graphs. Finally, we discuss the necessity of coarse-grained attribution analysis when studying long-context failure cases.

\subsection{Contrastive Attribution via LRP}

\paragraph{Model.}
We consider transformer-based autoregressive LLMs~\cite{NIPS2017_3f5ee243}.
Let $\mathcal{V}$ denote the vocabulary with $|\mathcal{V}|$ tokens and let $d$ be the hidden dimension.
The model comprises an embedding layer with matrix
$\mathbf{W}_E \in \mathbb{R}^{|\mathcal{V}| \times d}$,
followed by $L$ transformer blocks indexed by $l$.
%Each block consists of a multi-head self-attention (MHA) sublayer and a multi-layer perceptron (MLP) sublayer.
An unembedding matrix
$\mathbf{W}_U \in \mathbb{R}^{d \times |\mathcal{V}|}$
maps final hidden representations to vocabulary logits.

Given an input token sequence $\mathbf{t}_{\le i}=(t_0,\dots,t_i)$ with $t_j\in\mathcal{V}$,
the embedding layer produces token embeddings
$\mathbf{h}_{-1}=\mathbf{W}_E[\mathbf{t}_{\le i}]$.
Hidden states are propagated through the transformer stack as
\begin{equation}
\mathbf{h}_l=\mathrm{Transformer}_l(\mathbf{h}_{l-1}), \quad l=0,\dots,L-1.
\end{equation}
Let $\mathbf{h}_{L-1}^{(i)}\in\mathbb{R}^d$ denote the final-layer hidden state at position $i$.
The logits for next-token prediction are
\begin{equation}
\boldsymbol{\ell}_{i+1}
= \mathbf{h}_{L-1}^{(i)\top}\mathbf{W}_U \in \mathbb{R}^{|\mathcal{V}|}.
\end{equation}
% and the conditional probability distribution is
% \begin{equation}
% p(t_{i+1}\mid \mathbf{t}_{\le i})=\mathrm{softmax}(\boldsymbol{\ell}_{i+1}).
% \end{equation}

\paragraph{Contrastive attribution objective.}
We formulate failure analysis as a token-level \emph{contrastive attribution} problem.
Given an incorrect target token $t_{\mathrm{tgt}}$ and a contrast alternative $t_{\mathrm{con}}$,
we attribute the logit difference between them to input token embeddings and hidden states across layers.
The contrastive logit difference is
\begin{align}
\Delta \ell
&\coloneqq
\ell_{i+1}(t_{\mathrm{tgt}})-\ell_{i+1}(t_{\mathrm{con}}) \nonumber\\
&=
\mathbf{h}_{L-1}^{(i)\top}
\bigl(
\mathbf{W}_U[:,t_{\mathrm{tgt}}]-\mathbf{W}_U[:,t_{\mathrm{con}}]
\bigr),
\end{align}
where $\mathbf{W}_U[:,t]$ denotes the unembedding vector for token $t$.
This contrastive formulation closely reflects practical debugging scenarios (asking why the model prefers a specific incorrect token over a plausible alternative) and removes shared, failure-irrelevant computation, yielding more salient and interpretable explanations~\cite{DBLP:conf/emnlp/YinN22}.

\paragraph{Layer-wise Relevance Propagation.}
To attribute $\Delta \ell$ to intermediate representations, we employ Layer-wise Relevance Propagation (LRP)~\cite{bach2015pixel}.
LRP is an additive explanation method that decomposes a scalar quantity $R_j$ into contributions from input features
$\mathbf{x}=\{x_i\}_{i=1}^N$: $R_j=\sum_{i=1}^N R_{i\leftarrow j},$
where $R_{i\leftarrow j}$ denotes the relevance assigned to feature $x_i$ for explaining $R_j$.
%(note that $R_j$ may differ from the original function value by a constant scaling factor).
When a feature $i$ contributes to multiple downstream quantities $j$, its total relevance is obtained by aggregation $R_i=\sum_j R_{i\leftarrow j}$.

Applying LRP to our setting, we treat the contrastive logit difference $\Delta \ell$ as the attribution target and propagate relevance backward through all transformer layers to the embedding layer.
This yields a relevance vector $\mathbf{R}_i^{(l)}\in\mathbb{R}^d$ associated with each hidden state $\mathbf{h}_l^{(i)}$:
\begin{align}
\mathbf{R}_i^{(l)} &\leftrightarrow \mathbf{h}_l^{(i)}, \\
\mathbf{R}_i^{(l)} &= \sum_j \mathbf{R}_{\mathbf{h}_l^{(i)} \leftarrow \mathbf{h}_{l+1}^{(j)}},
\end{align}
where $\mathbf{R}_{\mathbf{h}_l^{(i)} \leftarrow \mathbf{h}_{l+1}^{(j)}}$ denotes the relevance propagated from hidden state $\mathbf{h}_{l+1}^{(j)}$ in the subsequent layer.
Collectively, these relevances define an \textbf{Attribution Graph}, whose nodes are $\mathbf{R}_i^{(l)}$ and whose edges correspond to propagated relevances.

\paragraph{LRP variant.}
Because transformers are highly non-linear, LRP relies on module-specific propagation rules that locally linearize model components.
Different rules yield different LRP variants.
As discussed earlier, although AttnLRP relaxes strict layer-wise relevance conservation, we adopt \emph{AttnLRP}~\cite{DBLP:conf/icml/AchtibatHDJWLS24} due to its superior attribution faithfulness compared to alternative attribution methods. A direct comparison in the context of failure diagnosis is provided in~\autoref{app:method_comparison}.

\subsection{Efficient Attribution Graph Construction}

Recent work shows that the relevance of input embeddings and hidden states under LRP can be efficiently computed via a modified \emph{gradient}$\times$\emph{input} formulation, enabling a single backward pass using standard automatic differentiation frameworks~\cite{DBLP:conf/icml/AchtibatHDJWLS24, arras2025closelookdecompositionbasedxaimethods}.
While this approach yields relevance scores for hidden states across layers, it does not directly provide the \emph{propagation structure} required to construct an attribution graph, i.e., how relevance flows between hidden states across layers.

Naively constructing the attribution graph is computationally prohibitive.
To recover relevance propagation details, one would need to treat each hidden-state relevance in layer $l+1$ as a separate attribution target and backpropagate it to layer $l$.
Given the large cardinality of $\{\mathbf{R}_i^{(l)}\}$, this results in an excessive number of backward passes.

To address this challenge, we leverage a batching trick that reuses the batch dimension to pack multiple attribution targets into a single backward pass, following recent work on attribution graph construction with sparse features in transcoders~\cite{hanna-etal-2025-circuit}.
This approach exploits GPU vectorization to efficiently recover relevance propagation between layers.
Details are provided in~\autoref{app:batch_packed_backprop}, with empirical efficiency gains reported in~\autoref{app:efficiency}.
To further improve graph interpretability, we prune attribution targets with relevance below a fixed threshold and remove edges with negligible relevance during graph construction.
The graph pruning strategy is described in~\autoref{app:graph_pruning}.

\subsection{Coarse-to-Fine Attribution Analysis}

The above formulation yields relevance at the level of individual hidden-state components, enabling neuron-level analysis~\cite{DBLP:conf/icml/AchtibatHDJWLS24}.
In practice, we also often aggregate relevance at the hidden-state level by summing over hidden state dimensions, $R_i^{(l)} = \sum_{k=1}^d \mathbf{R}_{i,k}^{(l)}$,
where $\mathbf{R}_{i,k}^{(l)}$ denotes the $k$-th component of $\mathbf{R}_i^{(l)}$.
We use this aggregated relevance to analyze input-token attribution patterns, e.g., via heatmaps (as shown in~\autoref{fig:URT_OIT}).

A coarse-grained treatment is also essential for attribution graph analysis.
Constructing attribution graphs is computationally expensive, particularly for long prompts.
To scale LRP-based interpretability to large collections of failure cases, we first construct attribution graphs at the hidden-state level by propagating aggregated relevance.
After identifying important subgraphs, we optionally refine them by propagating fine-grained, neuron-level relevance within the selected subgraph. This coarse-to-fine strategy differs from prior attribution graph approaches that directly operate at the neuron level (e.g.,~\cite{ameisen2025circuit}).
For long-context failure cases, a one-shot neuron-level construction is often computationally infeasible and hinders interpretability.

As an initial step, in this work we focus on the coarse-grained state-level analysis, and leave systematic neuron-level attribution graph analysis to future work.

\section{Experiments}

%In this section, we first describe the experimental setup, followed by three analyses: (i) characterizing how models prefer incorrect tokens across benchmarks and assess the extent to which contrastive attribution can explain these failure cases, (ii) examining how such failures are corrected by larger models, and (iii) analyzing how attribution patterns evolve across different training stages.

\subsection{Experimental Setup}

\begin{table*}[t]
\footnotesize
\centering
\caption{Summary of benchmarks and models, number of original failure cases, proportion of clean cases used for attribution analysis, and average input token count. Models highlighted in \textbf{bold} are used for failure trace generation.
%For Olmo3-7B-Think-SFT, we use the \texttt{step1000} checkpoint for failure collection. 
}
\label{tab:datasets}
\begin{tabular}{c | c | c |  c | c}
\hline
Benchmark & Target Capability & Model & Failure Cases (Clean Case Rate) & Avg. Token Count \\
\hline
\multirow{1}{*}{IFEval} &
\multirow{1}{*}{Instruction following} &
Qwen3-\textbf{0.6B}/1.7B/4B & 265 (20.8\%) & 54\\
% \cline{3-5}
%  &
%  &
% Olmo3-7B-Think-\textbf{SFT}/DPO/RLVR & - & -
% % & &
% \\
% \hline

GAIA2 &
\makecell[c]{Agentic; long-context} &
Qwen3-\textbf{4B} & 300 (17.0\%) &  12374
\\
% \hline

MATH &
Math &
Qwen3-\textbf{0.6B} & 91 (40.7\%) & 116
\\
% \hline

EvalPlus &
Coding &
Qwen3-\textbf{0.6B} & 270 (19.6\%) & 169
\\
\hline
\end{tabular}
\end{table*}

\paragraph{Benchmarks.}
To ensure our benchmark choices align with widely used ones while remaining relevant amid rapidly improving model capabilities, we follow the latest recommendations from HuggingFace~\cite{fourrier2025_the_llm_evaluation_guidebook}. Specifically, we adopt GAIA2~\cite{andrews2025arescalingagentenvironments} to represent long-horizon agentic tasks, IFEval~\cite{zhou2023instructionfollowingevaluationlargelanguage} to evaluate instruction-following, MATH~\cite{hendrycksmath2021} for math reasoning, and EvalPlus~\cite{evalplus, evalperf} for code generation.\footnote{We do not include additional agent-based coding benchmarks, as GAIA2 already captures agentic behavior; instead, we select EvalPlus to emphasize code-centric evaluation.} 
Together, these benchmarks span diverse domains, context lengths (e.g., 10k+ tokens for GAIA2 traces), and task types (agentic vs.\ non-agentic).

\paragraph{Models.}
We primarily use the open-source Qwen3 model series~\cite{yang2025qwen3technicalreport} for attribution analyses, and the Olmo-3-7B-Think model series~\cite{olmo2025olmo3} to study the evolution of attribution patterns across training checkpoints. Considering our computation resource constraints, we restrict the model sizes to below 8B. Benchmark-specific model configurations are summarized in~\autoref{tab:datasets}.

\paragraph{Failure Case Collection.}

To generate failure cases for analysis, we employed different model configuration across benchmarks, as summarized in~\autoref{tab:datasets}.
%\footnote{We use Qwen3-4B for GAIA2 instead of Qwen3-0.6B, as the latter often fails to produce coherent or goal-directed reasoning in this long-horizon setting, resulting in trivial failures.} 
Greedy decoding is employed for reproducibility. To ensure balanced analysis across datasets, we evaluated the full EvalPlus set due to its small size, while randomly sampling a subset from larger datasets IFEval, GAIA2, and MATH. More details on failure case collection can be found in~\autoref{app:failure_case_collection}. The number of collected failure cases per benchmark is also shown in~\autoref{tab:datasets}.

\paragraph{Contrast Token Pair Identification.}

After collecting failure cases, we apply a post-processing step to identify contrast token pairs for attribution analysis. This step can be challenging for several reasons: (i) there may be multiple plausible candidates, or none at all, for the target error token; and (ii) multiple choices may exist for the corresponding contrast token.

We define the \emph{target token} as the \textbf{earliest generated token} at which the model departs from a correct trajectory, following prior agent-failure attribution work~\citep{whoandwhen} which likewise localizes failure at the earliest error step. To identify this token reliably, we employ a two-stage pipeline: (1)~four independent LLMs each propose the target token, and (2)~human annotators validate cases with majority ($\geq$2) agreement among LLMs. Inter-agreement among LLM proposers under a relaxed $\pm 3$-token window exceeds 86\% across all benchmarks, and human approval of majority-agreed cases also exceeds 86\%, confirming that the identification is reliable (see~\autoref{app:target_token_agreement} for details).

The \emph{contrast token} is selected as the highest-ranked alternative during top-$k$ rollout, using both the models that generated the failure traces and stronger auxiliary models. Crucially, we impose a \textbf{strict recovery criterion}: the contrast token must, when substituted at the target position, actually \emph{recover} the correct trajectory. While this requirement could be relaxed by allowing alternative contrast tokens (even if the underlying model cannot recover from the failure due to limited capability), we enforce it to obtain clean, high-confidence contrastive pairs. Further discussion of this design choice, along with additional experiments using alternative contrast tokens, is provided in~\autoref{app:alternative_contrast_token_pairs}.

Finally, we impose a logit-difference threshold $\Delta \ell > 1$ between the target and contrast tokens. This threshold serves two purposes: (1)~it guards against numerical instability arising from finite-precision effects (e.g., int8 vs.\ FP16 inference can induce logit discrepancies up to ${\sim}0.5$); and (2)~a logit difference of 1 corresponds to an odds ratio of $e^1 \approx 2.7$, ensuring a meaningfully strong preference of the target token over the contrast token. 
%In practice, this threshold excludes only a small fraction of cases (\autoref{tab:filtering_breakdown}).

Using the above procedure, we identify high-quality contrast token pairs for \num{20.8}\% of failure cases for IFEval, \num{17.0}\% for GAIA2, \num{40.7}\% for MATH, and \num{19.6}\% for EvalPlus. Additional details on contrast token pair identification are provided in~\autoref{app:token_pair_identification}.

\paragraph{Attribution Analysis.}
We conduct attribution analysis in two stages.
First, we apply input attribution using AttnLRP, producing heatmaps that indicate each input token’s positive or negative contribution to the logit difference, as illustrated in~\autoref{fig:URT_OIT}.
Tokens with positive relevance (red) support the target token, whereas tokens with negative relevance (blue) disfavor the target token and instead support the contrast token.
Because the beginning-of-sequence (BOS) and other special tokens often receive disproportionately large relevance due to the attention sink effect~\cite{xiao2024efficient,cancedda2024spectral}, we exclude these special tokens (shown in gray in~\autoref{fig:URT_OIT}) when normalizing relevance scores.
Specifically, we normalize by the absolute value of the maximum relevance score among non-special tokens.

Second, we construct attribution graphs using our proposed batch-packed multi-target backpropagation method.
During graph construction, we discard nodes with absolute relevance below $0.01$ and prune edges using a threshold of $0.85$ under the per-layer cumulative mass pruning mode.

With the input attribution heatmaps and the attribution graphs, we perform a detailed manual analysis of identified clean failure cases.
The findings are independently verified by two authors; inter-annotator agreement statistics are reported in~\autoref{app:annotator_agreement}.
We leave the development of a fully automated attribution analysis pipeline to future work.

\subsection{Attribution Analysis of Incorrect Token Preference}

We first present attribution analysis results across four benchmarks, using the same model employed for failure trace generation. Our goal is to assess the extent to which attribution analysis can reveal explanatory signals for why the target token is preferred over its contrast token.

To this end, we categorize attribution outcomes as follows:
\begin{itemize}[leftmargin=*]
    \item \textbf{Manifested by Input Attribution (M-IA):} The failure is clearly observable in the input attribution heatmap.
    \item \textbf{No Clue from Input Attribution (NC-IA) + Manifested by Attribution Graph (M-AG):} The failure is not evident from input attribution alone but becomes apparent through attribution graph analysis.
    \item \textbf{No Clue from Both Input Attribution and Attribution Graph (NC-IA+AG):} The failure is not explained by either analysis, demanding finer-grain analysis.
\end{itemize}

\begin{figure}[htbp]
    \centering
    \includegraphics[width=0.7\columnwidth]{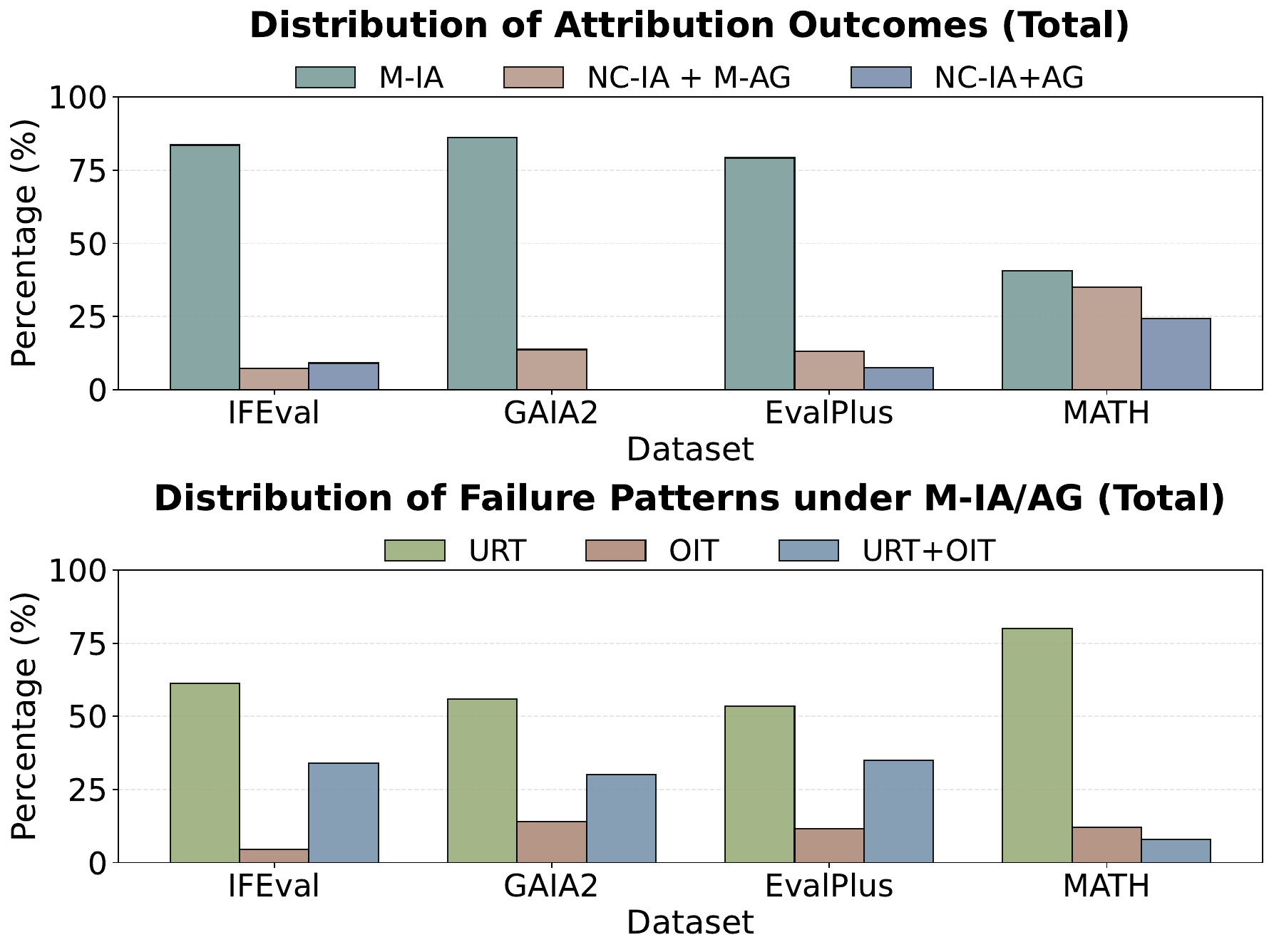}
    \caption{Distribution of attribution outcomes (top) and failure patterns (bottom) across benchmarks.}
    \label{fig:eval}
\end{figure}

The resulting categorization for the four benchmarks is shown in the top panel of~\autoref{fig:eval}. IFEval, GAIA2, and EvalPlus exhibit similar patterns: in the majority of cases, input attribution alone suffices to reveal failure signatures. In contrast, for MATH, a substantial fraction of failures remain unexplained even after attribution graph analysis. This limitation arises because our current attribution graphs operate at the level of aggregated hidden states, whereas numerical reasoning errors often require finer-grained, neuron-level attribution analysis~\cite{lindsey2025biology}. For example, for the case \textit{``Algebra\_9''} in MATH, Qwen3-0.6B incorrectly prefers the digit ``0'' over ``1'' following the prior context of ``$(1.0175)^{20} \approx 1.4$''. Explaining this error may require neuron-level tracing of the underlying power computation.

\begin{figure}[htbp]
    \centering
    \includegraphics[width=\columnwidth]{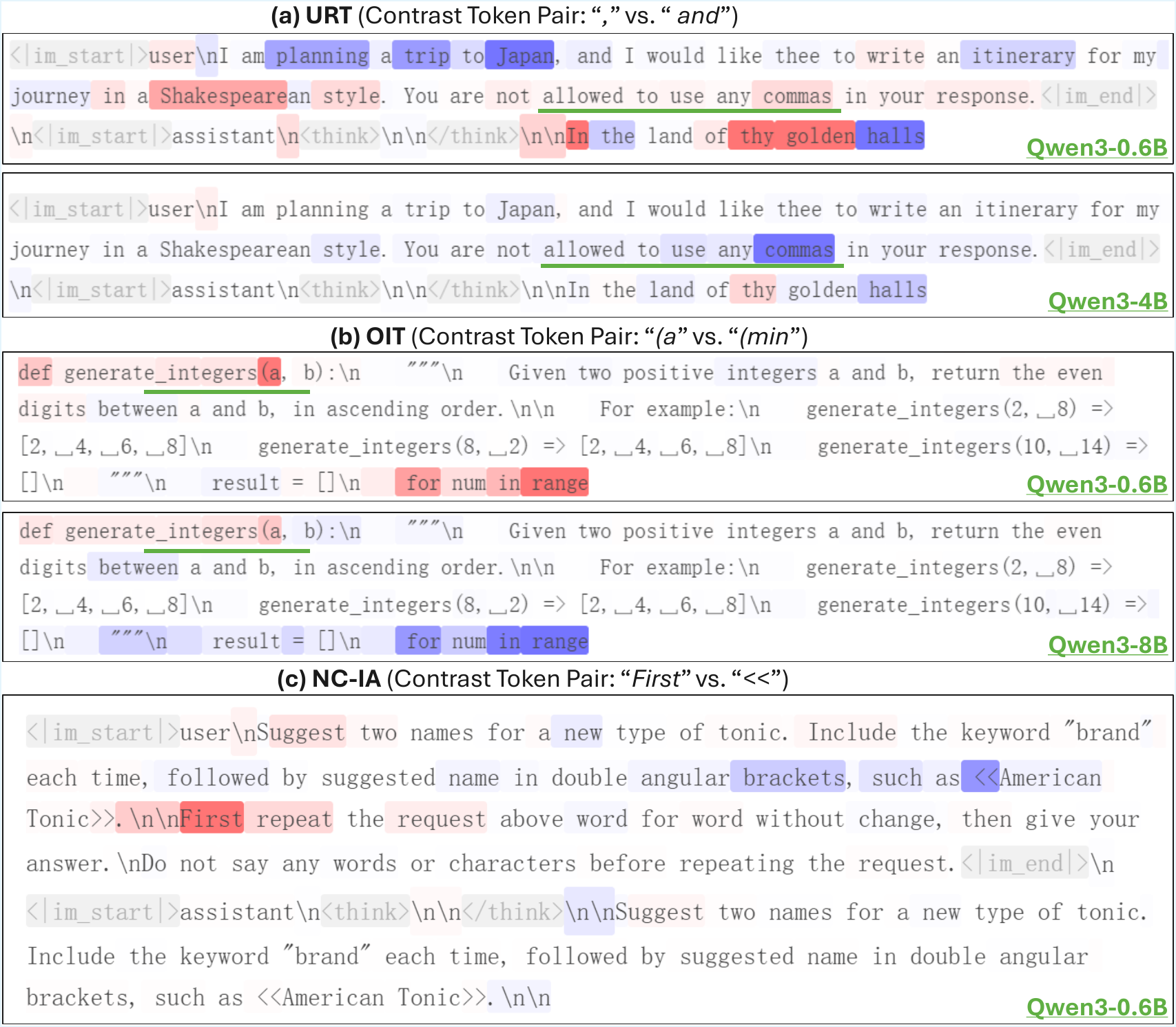}
    \caption{Examples of input attribution heatmaps. (a) URT: Qwen3-0.6B underweights the instruction token ``commas''. (b) OIT: Qwen3-0.6B overweights an irrelevant token ``(a''. (c) NC-IA: input attribution alone offers limited explanation, motivating attribution graph analysis. Color intensity shows each token’s positive (red) or negative (blue) influence on the target–contrast preference.}
    \label{fig:URT_OIT}
\end{figure}

For cases that are explainable by attribution analysis (i.e., those categorized as M-IA and NC-IA + M-AG), we further identify the following failure patterns:
\begin{itemize}[leftmargin=*]
    \item \textbf{Underweight Relevant Tokens (URT):} The model assigns insufficient relevance to critical input tokens. \autoref{fig:URT_OIT}(a) shows an example from IFEval, where Qwen3-0.6B fails to assign negative relevance to the keyword ``comma'', which is essential for producing the correct output. In contrast, Qwen3-4B assigns clear negative relevance to this token, aligning with correct behavior.
    \item \textbf{Overweight Irrelevant Tokens (OIT):} The model overemphasizes misleading or semantically uninformative tokens. Note that URT does not imply OIT, as AttnLRP does not enforce relevance conservation. As illustrated in \autoref{fig:URT_OIT}(b), a EvalPlus example shows Qwen3-0.6B assigning high relevance to an irrelevant token ``(a'', while much less contribution from that token in Qwen3-8B.
    \item \textbf{URT + OIT:} Failures characterized by both underweighting relevant tokens and overweighting irrelevant ones.
\end{itemize}

A fine-grained breakdown of the above failure patterns is shown in the bottom panel of~\autoref{fig:eval}. Across all four benchmarks, underweighting relevant tokens emerges as the dominant failure mode. Overweighting irrelevant tokens also contributes substantially, particularly for EvalPlus and GAIA2, and is often accompanied by underweighting of relevant tokens (except in MATH).

% \begin{figure}[htbp]
%     \centering
%     \begin{subfigure}[t]{\columnwidth}
%         \centering
%         \includegraphics[width=\linewidth]{figures/AG_CP_4_RE_exp.pdf}
%         \caption{NC-IA + M-AG example: relevance accumulates in later layers due to model-internal bias.}
%         \label{fig:attr_graph_a}
%     \end{subfigure}
%     \vspace{0.5em}
%     \begin{subfigure}[t]{\columnwidth}
%         \centering
%         \includegraphics[width=\linewidth]{figures/AG_no_clue_exp.pdf}
%         \caption{NC-IA + AG example: relevance dispersed across semantically unrelated tokens.}
%         \label{fig:attr_graph_b}
%     \end{subfigure}
%     \caption{Attribution graph analysis for two failure cases.\jz{let us discuss these two examples.}}
%     \label{fig:attr_graph_examples}
% \end{figure}
\begin{figure}[htbp]
    \centering
    \includegraphics[width=\columnwidth]{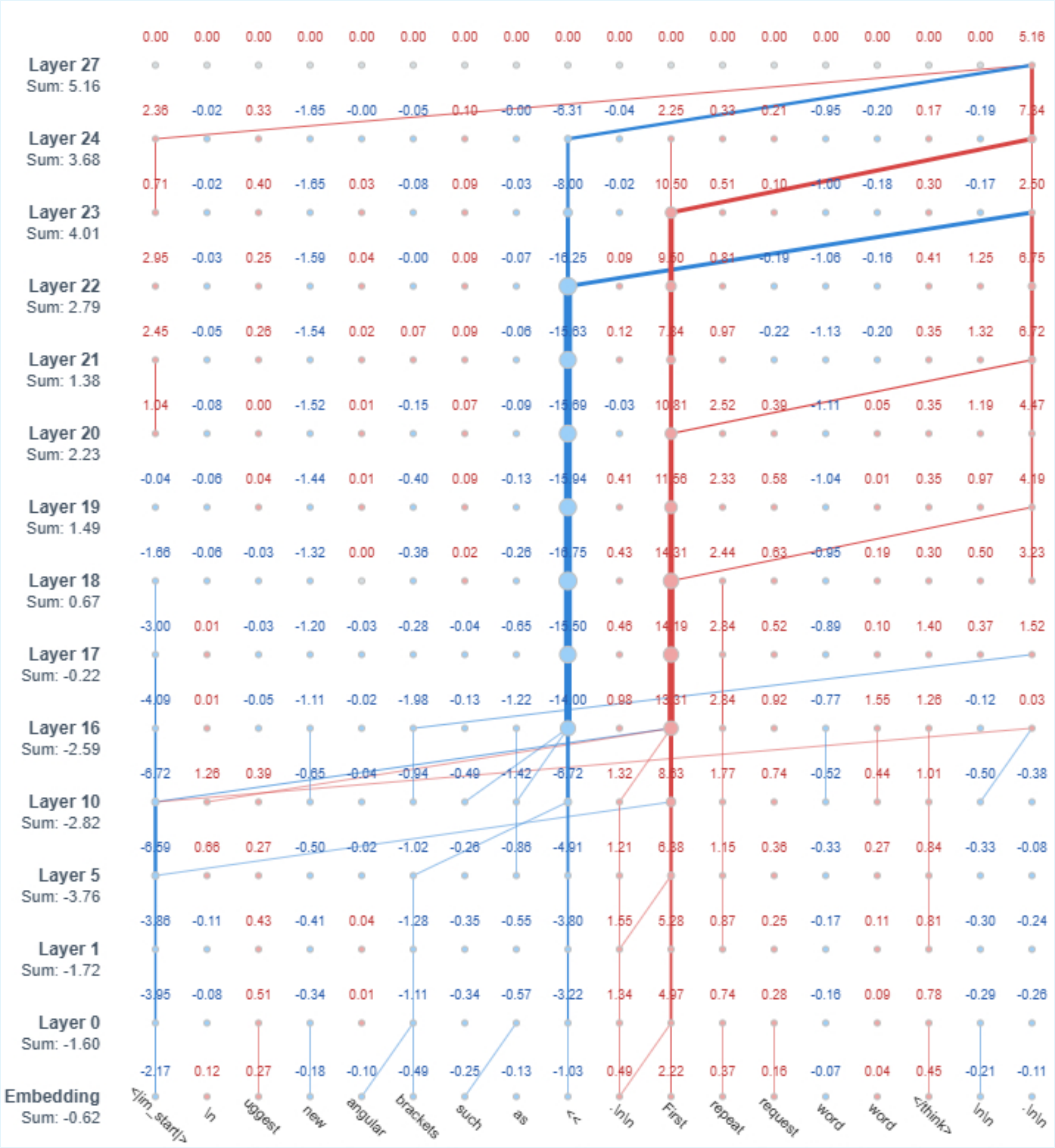}
    \caption{Sample ablated attribution graph for an NC-IA + M-AG case; the expanded attribution graph is provided in~\autoref{fig:fullag}.}
    \label{fig:AG}
\end{figure}

Lastly, we present a representative failure case in which input attribution offers limited insight, whereas the attribution graph reveals informative internal dynamics. As shown in~\autoref{fig:URT_OIT}(c), the output already satisfies the instruction to repeat the request and should therefore begin with the token ``\texttt{<<}''. Instead, the model erroneously outputs the token ``First''. Although the key tokens ``\texttt{<<}'' and ``brackets'' exhibit substantial negative relevance relative to the large positive relevance of ``First'', the model still yields a large final logit difference of $5.25$.
%this contrast might suggest only a marginal preference between the final competing tokens. However, the observed logit difference is $5.25$, indicating a clear preference against the target token.

To further analyze this behavior, we examine the attribution graph in~\autoref{fig:AG}, which highlights key tokens, layers, and relevance propagation paths. As relevance propagates to higher layers, the competing tokens ``First'' and ``\texttt{<<}'' initially show similar growth rates. At Layer~16, the absolute relevance of ``\texttt{<<}'' increases sharply (from $6.72$ to $14.0$), even surpassing that of ``First'', due to aggregated relevance from earlier tokens such as ``brackets such as''. Beyond Layer~16, ``First'' contributes several substantial increments to the residual stream at the final token position (Layers~18, 20, and~23), whereas ``\texttt{<<}'' makes two large contributions (Layers~22 and~24). The cumulative magnitude of relevance transferred by ``First'' ultimately exceeds that of ``\texttt{<<}'', resulting in a positive final logit difference and the observed erroneous output. Overall, this case illustrates how attribution graphs expose layer-wise relevance interactions that are invisible to input-level attribution, providing a more fine grained explanation of the model’s final decision. More example attribution graphs are given in~\autoref{app:AG}, and a systematic statistical analysis of attribution graph structure across all failure cases is provided in~\autoref{app:ag_statistics}.

\subsection{Attribution Shifts with Model Scaling}

We next investigate whether scaling model size can resolve the failures observed in smaller models and, if so, whether such improvements are supported by interpretability evidence. Aligning performance gains with interpretable mechanisms is crucial to verify that scaling improves models in the intended direction rather than exploiting spurious cues.

\begin{figure}[htbp]
    \centering
    \includegraphics[width=0.8\columnwidth]{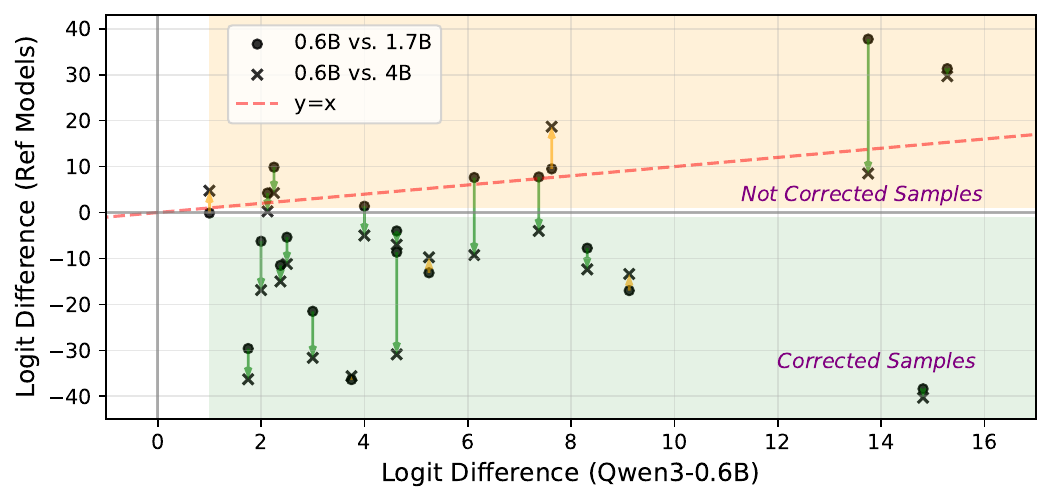}
    \caption{Logit difference comparisons between Qwen3-0.6B and larger models (1.7B and 4B) on IFEval failure cases. Most points lie below the $y=x$ line, indicating that larger models tend to correct failures in the smaller model. Green regions denote corrected samples, while yellow regions indicate remaining failures.}
    \label{fig:weak_strong_logit_diff}
\end{figure}

To this end, we apply larger Qwen3 models (1.7B and 4B) to the failure cases on IFEval identified for Qwen3-0.6B in the previous subsection. As shown in~\autoref{fig:weak_strong_logit_diff}, larger models consistently yield more negative logit differences for the contrast token pairs. In the plots comparing ``0.6B vs.\ 1.7B'' and ``0.6B vs.\ 4B'', most data points lie below the $y=x$ line, indicating that larger models tend to correct incorrect token preferences made by the smaller model. A similar trend is observed when comparing 1.7B and 4B, where many points exhibit further negative shifts (green downward arrows). While the majority of failures are corrected, as reflected by the concentration of points in the green shaded region, a subset of cases remains uncorrected (yellow region).

\begin{figure}[htbp]
    \centering
    \includegraphics[width=0.8\columnwidth]{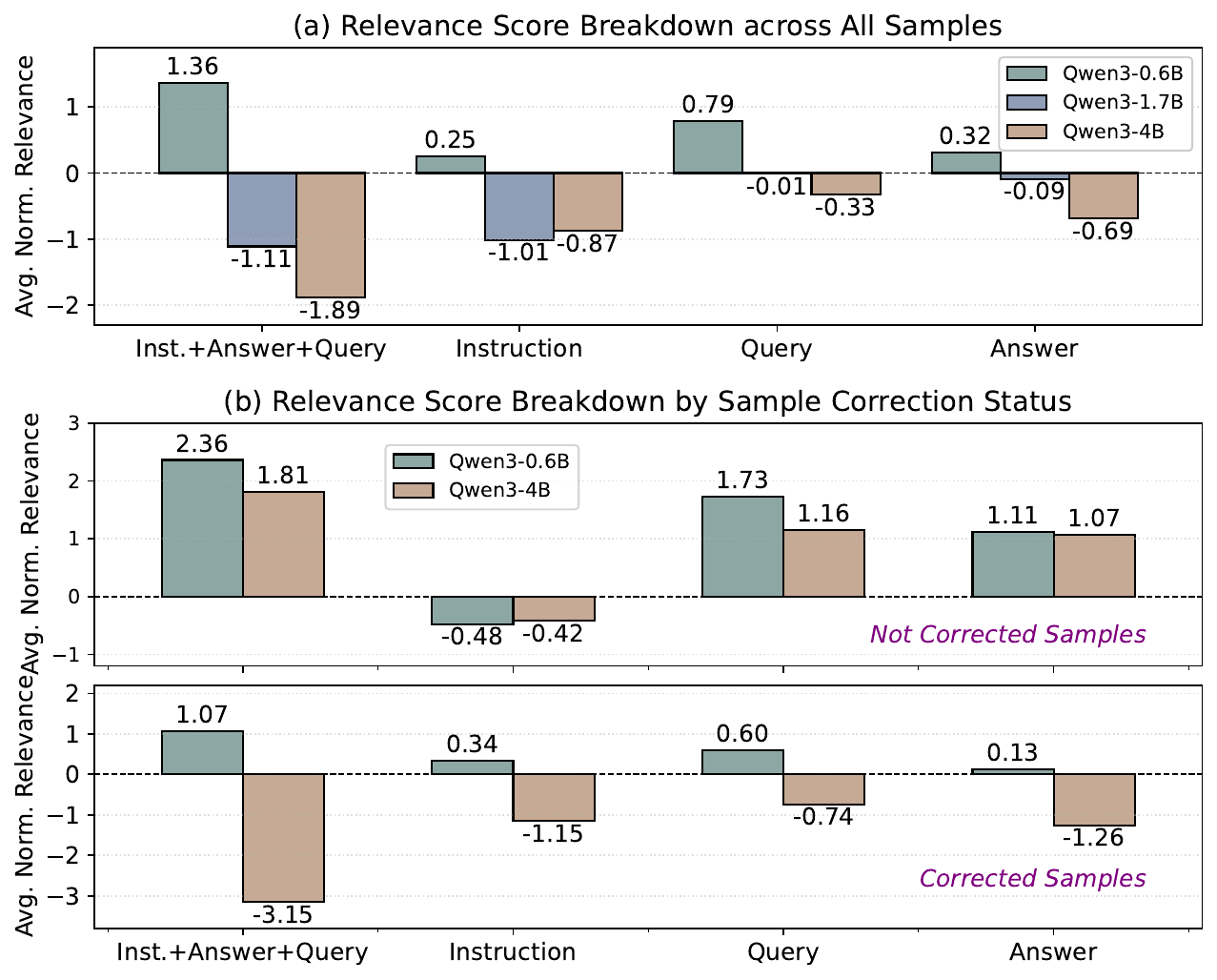}
    \caption{Input attribution relevance score breakdown across prompt segments. \textbf{(a)} Average normalized relevance over all samples. \textbf{(b)} Relevance breakdown by correction status.}
    \label{fig:weak_strong_rel_breakdown}
\end{figure}

We next examine whether these improvements are supported by interpretability analyses. Specifically, we perform input attribution for all three models on the same cases and compute normalized relevance score distributions over three prompt segments: \emph{Instruction} (constraint-related tokens, e.g., ``not allowed to use any commas''), \emph{Query} (the task description), and \emph{Answer} (tokens after \texttt{</think>}). The results are shown in~\autoref{fig:weak_strong_rel_breakdown}. The top panel demonstrates that total relevance scores become increasingly negative as model size increases from 0.6B to 1.7B and 4B. Notably, the Instruction segment, which exhibits non-negative relevance in the 0.6B model, becomes clearly negative in the larger models. In addition, both Query and Answer segments also show more negative relevance, suggesting that larger models are more attentive to task requirements and less driven by superficial token continuation from prior answer tokens.

This pattern becomes more pronounced when separating corrected and uncorrected samples, as shown in the bottom panel of~\autoref{fig:weak_strong_rel_breakdown}. For corrected samples, all three segments exhibit negative relevance scores, whereas uncorrected samples display nearly identical relevance distributions across model sizes. These findings indicate that failure correction in larger models is associated with systematic and interpretable shifts in attribution patterns. Overall, the results provide evidence that scaling model size leads to genuine improvements and strengthens confidence in model scaling as a principled approach to enhancing performance. 
%Additional qualitative comparisons of input attribution heatmaps for representative cases are provided in~\autoref{fig:URT_OIT}.

\subsection{Evolution of Failure Attribution Across Training}

Beyond studying the model scaling effects, we further apply them to investigate how model behavior evolves over the course of training. Specifically, we analyze input attribution patterns across post-training checkpoints of the Olmo-3-7B-Think model series~\cite{olmo2025olmo3}.

\begin{figure}[htbp]
    \centering
    \includegraphics[width=0.8\columnwidth]{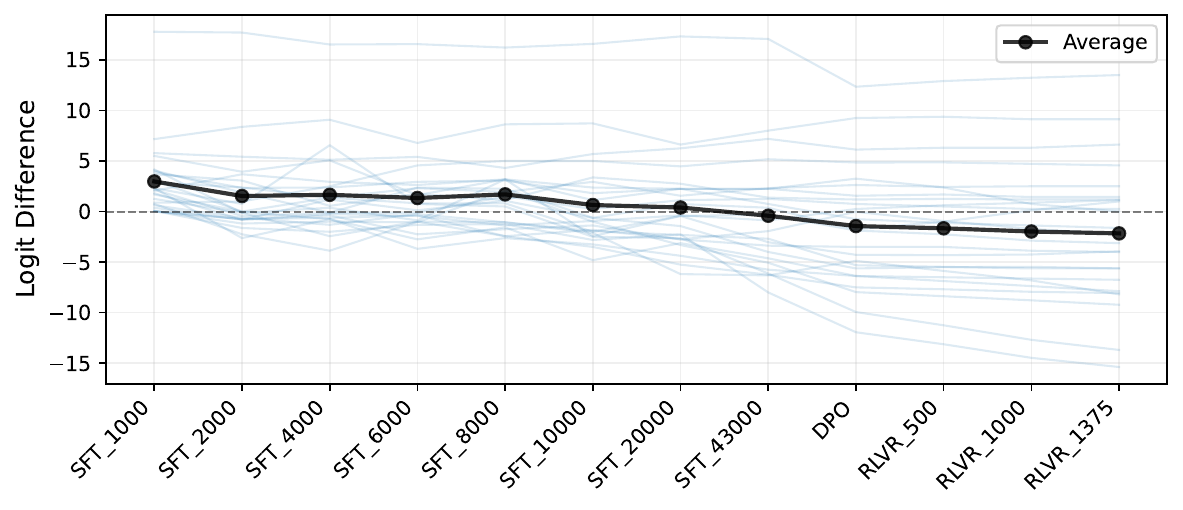}
    \caption{Evolution of logit differences across training checkpoints for Olmo-3-7B-Think on IFEval failure cases identified at \texttt{SFT\_1000}. Each line corresponds to an individual sample, with the bold line denoting the average. 
    %Logit differences steadily decrease as training progresses through SFT, Direct Preference Optimization (DPO), and Reinforcement Learning with Verifiable Rewards (RLVR), indicating gradual correction of earlier failures.
    }
    \label{fig:checkpoints_logit_diff}
\end{figure}

We begin by identifying failure samples on IFEval using an early checkpoint, \texttt{SFT\_1000}, at step 1000 of the supervised fine-tuning (SFT) stage. After extracting the corresponding contrast token pairs, we perform input attribution for these same cases across later checkpoints, including eight checkpoints during SFT, one checkpoint after Direct Preference Optimization (DPO), and three checkpoints in the final Reinforcement Learning with Verifiable Rewards (RLVR) stage. As shown in~\autoref{fig:checkpoints_logit_diff}, we observe a steady reduction in overall logit differences as training progresses. 
%The figure visualizes both the evolution trajectories of individual samples and their average trend, indicating that continued training gradually corrects earlier failures.

\begin{figure}[htbp]
    \centering
    \includegraphics[width=0.8\columnwidth]{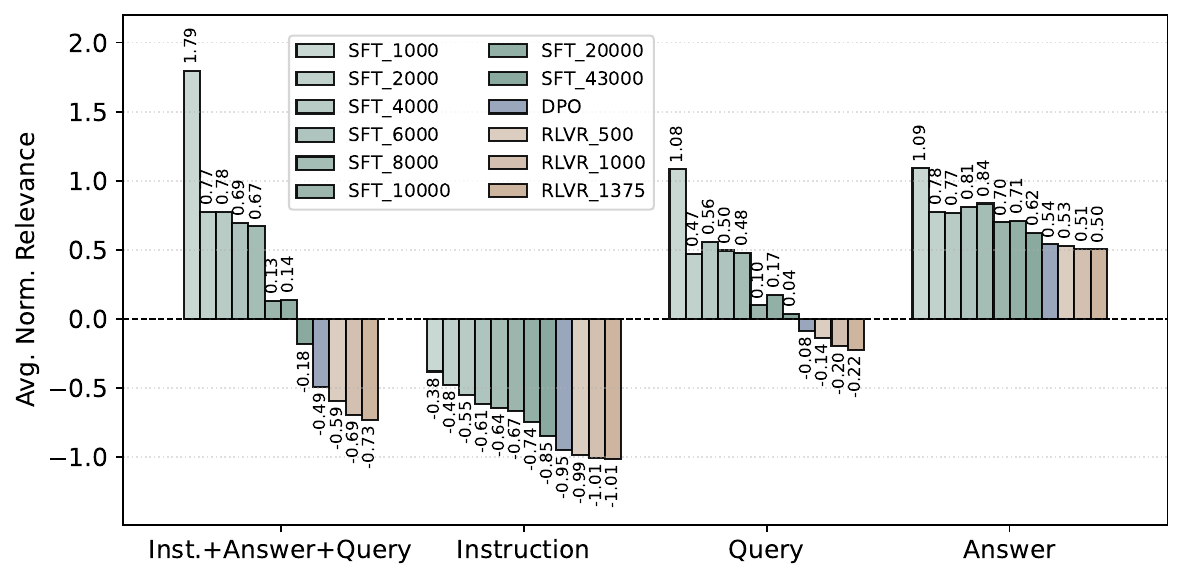}
    \caption{Input attribution relevance score breakdown across training checkpoints by prompt segment. Relevance is shown for \emph{Instruction}, \emph{Query}, and \emph{Answer} segments. 
    %The largest relevance shifts occur during early and mid SFT stages, primarily driven by reductions in Query and Answer relevance, while Instruction relevance decreases more steadily. Later stages such as DPO further push relevance across all segments toward more negative values.
    }
    \label{fig:checkpoints_relevance_breakdown}
\end{figure}

Analogous to our analysis of model scaling, we further examine how attribution patterns evolve across checkpoints by computing relevance score breakdowns over prompt segments, as shown in~\autoref{fig:checkpoints_relevance_breakdown}. Two key observations emerge. First, the most substantial relevance changes occur during the early SFT phase, particularly between \texttt{SFT\_1000} and \texttt{SFT\_2000}. This reduction is primarily driven by decreased relevance in the \emph{Query} and \emph{Answer} segments, suggesting that early SFT helps the model become more attentive to task semantics while reducing reliance on superficial token continuation from prior answers. A second notable relevance drop appears during the middle SFT stage, between \texttt{SFT\_8000} and \texttt{SFT\_10000}, again largely attributable to changes in the Query and Answer segments. Importantly, these inflection points align closely with the corresponding reductions in logit differences observed in~\autoref{fig:checkpoints_logit_diff}.

Second, while relevance in Query and Answer exhibits pronounced shifts, relevance associated with \emph{Instruction} decreases more steadily across checkpoints. Notably, comparing the final SFT checkpoint (\texttt{SFT\_43000}) with the post-DPO checkpoint reveals that DPO further pushes relevances in all three segments toward more negative values.

We note that the magnitude of attribution shifts differs across training stages: SFT induces the largest changes, while RLVR contributes more modestly. This discrepancy can be attributed to two factors: (1)~IFEval probes instruction-following capabilities that are more directly targeted by SFT than by RLVR, which primarily focuses on multi-step reasoning; and (2)~SFT precedes RLVR in the training pipeline, so the model may have already corrected many failure modes during SFT, leaving less room for RLVR to further shift attribution patterns.

Overall, these interpretability-based analyses provide fine-grained evidence that training progressively corrects failures through systematic attribution shifts. This highlights the utility of attribution-based diagnostics as a practical tool to monitor and debug model behavior throughout training.

\subsection{Practical Implications}

Beyond its diagnostic role, our attribution-based analysis suggests several actionable directions for model behavior improvement.

\paragraph{Targeted prompt tuning.}
Our perturbation experiments (see~\autoref{app:method_comparison}) show that masking a small number of top-attributed input tokens (${\sim}2$ on average) suffices to flip the model's prediction away from the error token. This suggests a practical approach for debugging and improving model behavior through targeted prompt adjustments: practitioners can use attribution heatmaps to identify which input tokens most strongly drive incorrect outputs and refine those specific prompt segments, greatly enhancing the efficiency of prompt engineering.

\paragraph{Monitoring model behavior during training.}
Our checkpoint-wise and model-scaling experiments (Sections~4.3 and~4.4) demonstrate that attribution patterns can track how failures evolve during training, serving as a diagnostic tool for model development. Such interpretability-based monitoring can reveal whether failure correction via model scaling or continued training is due to genuine changes in the underlying reasoning process (e.g., correctly weighting relevant tokens) rather than superficial changes (e.g., memorizing specific token patterns).

\paragraph{Token-level signals for alignment training.}
Attribution scores may provide a natural token-importance signal for alignment training. Instead of treating all tokens in a preferred or dispreferred response equally, one could use attribution to emphasize the tokens that actually drive the failure or preference gap. This suggests a natural integration with token-level preference optimization methods such as DPO~\cite{rafailov2023direct}. Recent work on Token-Importance Guided Direct Preference Optimization~\cite{TIDPO2026} similarly shows that token-importance weighting can improve alignment robustness and stability, though their approach relies on gradient-based token importance, which can be noisy and less faithful. A promising direction is to use more faithful attribution methods, such as LRP-based scores, to derive token-importance signals for alignment training.

\section{Conclusion}

We studied contrastive, LRP-based attribution as a practical tool for analyzing LLM failures on realistic benchmarks. By framing failure analysis as explaining why a model prefers an incorrect token over a correct alternative, and introducing an efficient method for constructing cross-layer attribution graphs, we applied interpretability at scales and settings largely unexplored by prior work. Across benchmarks, model sizes, and training checkpoints, we find that contrastive attribution can reveal meaningful failure patterns in many cases (e.g., underweighting relevant context) and that improvements from model scaling and continued training are often accompanied by systematic, interpretable shifts in attribution. Meanwhile, a non-trivial fraction of failures remain unexplained, particularly in math reasoning, highlighting limitations of our coarse grain attribution methods. Overall, our results suggest that interpretability can offer concrete diagnostic value for realistic LLM failure analysis, while also underscoring the need for more expressive and scalable tools to fully explain complex model errors. An important direction for future work is to extend the current token-level contrastive framework to handle multi-step failure modes driven by accumulative errors, for example by moving toward phrase-level or step-level contrastive attribution that can capture failures emerging from a sequence of reasoning steps rather than at a single identifiable token.

% \section*{Accessibility}

% Authors are kindly asked to make their submissions as accessible as possible
% for everyone including people with disabilities and sensory or neurological
% differences. Tips of how to achieve this and what to pay attention to will be
% provided on the conference website \url{http://icml.cc/}.
\section*{Acknowledgements}

We thank Haibin Lai for support with experimental annotation, 
and Bowen Zhang and Yangfan Qiao for helpful suggestions on our demo pipeline.

\clearpage

\bibliography{MIFA_ICML26}

@inproceedings{NIPS2017_3f5ee243,
 author = {Vaswani, Ashish and Shazeer, Noam and Parmar, Niki and Uszkoreit, Jakob and Jones, Llion and Gomez, Aidan N and Kaiser, \L ukasz and Polosukhin, Illia},
 booktitle = {Advances in Neural Information Processing Systems},
 editor = {I. Guyon and U. Von Luxburg and S. Bengio and H. Wallach and R. Fergus and S. Vishwanathan and R. Garnett},
 pages = {},
 publisher = {Curran Associates, Inc.},
 title = {Attention is All you Need},
 url_long = {https://proceedings.neurips.cc/paper_files/paper/2017/file/3f5ee243547dee91fbd053c1c4a845aa-Paper.pdf},
 volume = {30},
 year = {2017}
}

@article{eval_survey_1,
author = {Chang, Yupeng and Wang, Xu and Wang, Jindong and Wu, Yuan and Yang, Linyi and Zhu, Kaijie and Chen, Hao and Yi, Xiaoyuan and Wang, Cunxiang and Wang, Yidong and Ye, Wei and Zhang, Yue and Chang, Yi and Yu, Philip S. and Yang, Qiang and Xie, Xing},
title = {A Survey on Evaluation of Large Language Models},
year = {2024},
issue_date = {June 2024},
publisher = {Association for Computing Machinery},
address = {New York, NY, USA},
volume = {15},
number = {3},
issn = {2157-6904},
url = {https://doi.org/10.1145/3641289},
doi = {10.1145/3641289},
journal = {ACM Trans. Intell. Syst. Technol.},
month = mar,
articleno = {39},
numpages = {45}
}

@inproceedings{maynez-etal-2020-faithfulness,
    title = "On Faithfulness and Factuality in Abstractive Summarization",
    author = "Maynez, Joshua  and
      Narayan, Shashi  and
      Bohnet, Bernd  and
      McDonald, Ryan",
    editor = "Jurafsky, Dan  and
      Chai, Joyce  and
      Schluter, Natalie  and
      Tetreault, Joel",
    booktitle = "Proceedings of the 58th Annual Meeting of the Association for Computational Linguistics",
    month = jul,
    year = "2020",
    address = "Online",
    publisher = "Association for Computational Linguistics",
    url = "https://aclanthology.org/2020.acl-main.173/",
    doi = "10.18653/v1/2020.acl-main.173",
    pages = "1906--1919"
}

@article{hallu_survey_1,
author = {Ji, Ziwei and Lee, Nayeon and Frieske, Rita and Yu, Tiezheng and Su, Dan and Xu, Yan and Ishii, Etsuko and Bang, Ye Jin and Madotto, Andrea and Fung, Pascale},
title = {Survey of Hallucination in Natural Language Generation},
year = {2023},
issue_date = {December 2023},
publisher = {Association for Computing Machinery},
address = {New York, NY, USA},
volume = {55},
number = {12},
issn = {0360-0300},
url = {https://doi.org/10.1145/3571730},
doi = {10.1145/3571730},
journal = {ACM Comput. Surv.},
month = mar,
articleno = {248},
numpages = {38}
}

@article{hallu_survey_2,
author = {Huang, Lei and Yu, Weijiang and Ma, Weitao and Zhong, Weihong and Feng, Zhangyin and Wang, Haotian and Chen, Qianglong and Peng, Weihua and Feng, Xiaocheng and Qin, Bing and Liu, Ting},
title = {A Survey on Hallucination in Large Language Models: Principles, Taxonomy, Challenges, and Open Questions},
year = {2025},
issue_date = {March 2025},
publisher = {Association for Computing Machinery},
address = {New York, NY, USA},
volume = {43},
number = {2},
issn = {1046-8188},
url = {https://doi.org/10.1145/3703155},
doi = {10.1145/3703155},
journal = {ACM Trans. Inf. Syst.},
month = jan,
articleno = {42},
numpages = {55}
}

@article{hallu_survey_3,
    author = {Zhang, Yue and Li, Yafu and Cui, Leyang and Cai, Deng and Liu, Lemao and Fu, Tingchen and Huang, Xinting and Zhao, Enbo and Zhang, Yu and Chen, Yulong and Wang, Longyue and Luu, Anh Tuan and Bi, Wei and Shi, Freda and Shi, Shuming},
    title = {Siren’s Song in the AI Ocean: A Survey on Hallucination in Large Language Models},
    journal = {Computational Linguistics},
    volume = {51},
    number = {4},
    pages = {1373-1418},
    year = {2025},
    month = {12},
    issn = {0891-2017},
    doi = {10.1162/COLI.a.16},
    url = {https://doi.org/10.1162/COLI.a.16},
    eprint = {https://direct.mit.edu/coli/article-pdf/51/4/1373/2535477/coli.a.16.pdf},
}

@misc{fourrier2025_the_llm_evaluation_guidebook,
  title={The LLM Evaluation Guidebook},
  author={Clémentine Fourrier and Thibaud Frere and Guilherme Penedo and Thomas Wolf},
  year={2025},
  url = {https://huggingface.co/spaces/OpenEvals/evaluation-guidebook#recommendations}
  
}

@inproceedings{rome,
 author = {Meng, Kevin and Bau, David and Andonian, Alex and Belinkov, Yonatan},
 booktitle = {Advances in Neural Information Processing Systems},
 editor = {S. Koyejo and S. Mohamed and A. Agarwal and D. Belgrave and K. Cho and A. Oh},
 pages = {17359--17372},
 publisher = {Curran Associates, Inc.},
 title = {Locating and Editing Factual Associations in GPT},
 url_long = {https://proceedings.neurips.cc/paper_files/paper/2022/file/6f1d43d5a82a37e89b0665b33bf3a182-Paper-Conference.pdf},
 volume = {35},
 year = {2022}
}

@inproceedings{dai-etal-2022-knowledge,
    title = "Knowledge Neurons in Pretrained Transformers",
    author = "Dai, Damai  and
      Dong, Li  and
      Hao, Yaru  and
      Sui, Zhifang  and
      Chang, Baobao  and
      Wei, Furu",
    editor = "Muresan, Smaranda  and
      Nakov, Preslav  and
      Villavicencio, Aline",
    booktitle = "Proceedings of the 60th Annual Meeting of the Association for Computational Linguistics (Volume 1: Long Papers)",
    month = may,
    year = "2022",
    address = "Dublin, Ireland",
    publisher = "Association for Computational Linguistics",
    url = "https://aclanthology.org/2022.acl-long.581/",
    doi = "10.18653/v1/2022.acl-long.581",
    pages = "8493--8502"
}

@inproceedings{geva-etal-2021-transformer,
    title = "Transformer Feed-Forward Layers Are Key-Value Memories",
    author = "Geva, Mor  and
      Schuster, Roei  and
      Berant, Jonathan  and
      Levy, Omer",
    editor = "Moens, Marie-Francine  and
      Huang, Xuanjing  and
      Specia, Lucia  and
      Yih, Scott Wen-tau",
    booktitle = "Proceedings of the 2021 Conference on Empirical Methods in Natural Language Processing",
    month = nov,
    year = "2021",
    address = "Online and Punta Cana, Dominican Republic",
    publisher = "Association for Computational Linguistics",
    url = "https://aclanthology.org/2021.emnlp-main.446/",
    doi = "10.18653/v1/2021.emnlp-main.446",
    pages = "5484--5495"
}

@inproceedings{
meng2023massediting,
title={Mass-Editing Memory in a Transformer},
author={Kevin Meng and Arnab Sen Sharma and Alex J Andonian and Yonatan Belinkov and David Bau},
booktitle={The Eleventh International Conference on Learning Representations },
year={2023},
url={https://openreview.net/forum?id=MkbcAHIYgyS}
}

@inproceedings{NEURIPS2024_d6df31b1,
 author = {Yao, Yunzhi and Zhang, Ningyu and Xi, Zekun and Wang, Mengru and Xu, Ziwen and Deng, Shumin and Chen, Huajun},
 booktitle = {Advances in Neural Information Processing Systems},
 doi = {10.52202/079017-3765},
 editor = {A. Globerson and L. Mackey and D. Belgrave and A. Fan and U. Paquet and J. Tomczak and C. Zhang},
 pages = {118571--118602},
 publisher = {Curran Associates, Inc.},
 title = {Knowledge Circuits in Pretrained Transformers},
 url_long = {https://proceedings.neurips.cc/paper_files/paper/2024/file/d6df31b1be98e04be48af8bedb95b499-Paper-Conference.pdf},
 volume = {37},
 year = {2024}
}

@inproceedings{trufulQA,
    title = "{T}ruthful{QA}: Measuring How Models Mimic Human Falsehoods",
    author = "Lin, Stephanie  and
      Hilton, Jacob  and
      Evans, Owain",
    editor = "Muresan, Smaranda  and
      Nakov, Preslav  and
      Villavicencio, Aline",
    booktitle = "Proceedings of the 60th Annual Meeting of the Association for Computational Linguistics (Volume 1: Long Papers)",
    month = may,
    year = "2022",
    address = "Dublin, Ireland",
    publisher = "Association for Computational Linguistics",
    url = "https://aclanthology.org/2022.acl-long.229/",
    doi = "10.18653/v1/2022.acl-long.229",
    pages = "3214--3252"
}

@inproceedings{HaluEval,
    title = "{H}alu{E}val: A Large-Scale Hallucination Evaluation Benchmark for Large Language Models",
    author = "Li, Junyi  and
      Cheng, Xiaoxue  and
      Zhao, Xin  and
      Nie, Jian-Yun  and
      Wen, Ji-Rong",
    editor = "Bouamor, Houda  and
      Pino, Juan  and
      Bali, Kalika",
    booktitle = "Proceedings of the 2023 Conference on Empirical Methods in Natural Language Processing",
    month = dec,
    year = "2023",
    address = "Singapore",
    publisher = "Association for Computational Linguistics",
    url = "https://aclanthology.org/2023.emnlp-main.397/",
    doi = "10.18653/v1/2023.emnlp-main.397",
    pages = "6449--6464"
}

@misc{HalluLens,
      title={HalluLens: LLM Hallucination Benchmark}, 
      author={Yejin Bang and Ziwei Ji and Alan Schelten and Anthony Hartshorn and Tara Fowler and Cheng Zhang and Nicola Cancedda and Pascale Fung},
      year={2025},
      eprint={2504.17550},
      archivePrefix={arXiv},
      primaryClass={cs.CL},
      url={https://arxiv.org/abs/2504.17550}, 
}

@misc{luo2024understandingutilizationsurveyexplainability,
      title={From Understanding to Utilization: A Survey on Explainability for Large Language Models}, 
      author={Haoyan Luo and Lucia Specia},
      year={2024},
      eprint={2401.12874},
      archivePrefix={arXiv},
      primaryClass={cs.CL},
      url={https://arxiv.org/abs/2401.12874}, 
}

@inproceedings{
song2025a,
title={A Survey on Large Language Model Reasoning Failures},
author={Peiyang Song and Pengrui Han and Noah Goodman},
booktitle={2nd AI for Math Workshop @ ICML 2025},
year={2025},
url={https://openreview.net/forum?id=hsgMn4KBFG}
}

@misc{li202512surveyreasoning,
      title={From System 1 to System 2: A Survey of Reasoning Large Language Models}, 
      author={Zhong-Zhi Li and Duzhen Zhang and Ming-Liang Zhang and Jiaxin Zhang and Zengyan Liu and Yuxuan Yao and Haotian Xu and Junhao Zheng and Pei-Jie Wang and Xiuyi Chen and Yingying Zhang and Fei Yin and Jiahua Dong and Zhiwei Li and Bao-Long Bi and Ling-Rui Mei and Junfeng Fang and Xiao Liang and Zhijiang Guo and Le Song and Cheng-Lin Liu},
      year={2025},
      eprint={2502.17419},
      archivePrefix={arXiv},
      primaryClass={cs.AI},
      url={https://arxiv.org/abs/2502.17419}, 
}

@misc{baker2025monitoring,
      title={Monitoring Reasoning Models for Misbehavior and the Risks of Promoting Obfuscation}, 
      author={Bowen Baker and Joost Huizinga and Leo Gao and Zehao Dou and Melody Y. Guan and Aleksander Madry and Wojciech Zaremba and Jakub Pachocki and David Farhi},
      year={2025},
      eprint={2503.11926},
      archivePrefix={arXiv},
      primaryClass={cs.AI},
      url={https://arxiv.org/abs/2503.11926}, 
}

@misc{galichin2025icoveredbaseshere,
      title={I Have Covered All the Bases Here: Interpreting Reasoning Features in Large Language Models via Sparse Autoencoders}, 
      author={Andrey Galichin and Alexey Dontsov and Polina Druzhinina and Anton Razzhigaev and Oleg Y. Rogov and Elena Tutubalina and Ivan Oseledets},
      year={2025},
      eprint={2503.18878},
      archivePrefix={arXiv},
      primaryClass={cs.CL},
      url={https://arxiv.org/abs/2503.18878}, 
}

@inproceedings{zhang-etal-2025-reasoning,
    title = "From Reasoning to Answer: Empirical, Attention-Based and Mechanistic Insights into Distilled {D}eep{S}eek R1 Models",
    author = "Zhang, Jue  and
      Lin, Qingwei  and
      Rajmohan, Saravan  and
      Zhang, Dongmei",
    editor = "Christodoulopoulos, Christos  and
      Chakraborty, Tanmoy  and
      Rose, Carolyn  and
      Peng, Violet",
    booktitle = "Proceedings of the 2025 Conference on Empirical Methods in Natural Language Processing",
    month = nov,
    year = "2025",
    address = "Suzhou, China",
    publisher = "Association for Computational Linguistics",
    url = "https://aclanthology.org/2025.emnlp-main.198/",
    doi = "10.18653/v1/2025.emnlp-main.198",
    pages = "3985--4002",
    ISBN = "979-8-89176-332-6"
}

@misc{MAST,
      title={Why Do Multi-Agent LLM Systems Fail?}, 
      author={Mert Cemri and Melissa Z. Pan and Shuyi Yang and Lakshya A. Agrawal and Bhavya Chopra and Rishabh Tiwari and Kurt Keutzer and Aditya Parameswaran and Dan Klein and Kannan Ramchandran and Matei Zaharia and Joseph E. Gonzalez and Ion Stoica},
      year={2025},
      eprint={2503.13657},
      archivePrefix={arXiv},
      primaryClass={cs.AI},
      url={https://arxiv.org/abs/2503.13657}, 
}

@inproceedings{
whoandwhen,
title={Which Agent Causes Task Failures and When? On Automated Failure Attribution of {LLM} Multi-Agent Systems},
author={Shaokun Zhang and Ming Yin and Jieyu Zhang and Jiale Liu and Zhiguang Han and Jingyang Zhang and Beibin Li and Chi Wang and Huazheng Wang and Yiran Chen and Qingyun Wu},
booktitle={Forty-second International Conference on Machine Learning},
year={2025},
url={https://openreview.net/forum?id=GazlTYxZss}
}

@misc{MI_survey,
      title={A Primer on the Inner Workings of Transformer-based Language Models}, 
      author={Javier Ferrando and Gabriele Sarti and Arianna Bisazza and Marta R. Costa-jussà},
      year={2024},
      eprint={2405.00208},
      archivePrefix={arXiv},
      primaryClass={cs.CL},
      url={https://arxiv.org/abs/2405.00208}, 
}

@inproceedings{DBLP:journals/corr/SimonyanVZ13,
  author       = {Karen Simonyan and
                  Andrea Vedaldi and
                  Andrew Zisserman},
  editor       = {Yoshua Bengio and
                  Yann LeCun},
  title        = {Deep Inside Convolutional Networks: Visualising Image Classification
                  Models and Saliency Maps},
  booktitle    = {2nd International Conference on Learning Representations, {ICLR} 2014,
                  Banff, AB, Canada, April 14-16, 2014, Workshop Track Proceedings},
  year         = {2014},
  url          = {http://arxiv.org/abs/1312.6034},
  bibsource    = {dblp computer science bibliography, https://dblp.org}
}

@misc{denil2015extractionsalientsentenceslabelled,
      title={Extraction of Salient Sentences from Labelled Documents}, 
      author={Misha Denil and Alban Demiraj and Nando de Freitas},
      year={2015},
      eprint={1412.6815},
      archivePrefix={arXiv},
      primaryClass={cs.CL},
      url={https://arxiv.org/abs/1412.6815}, 
}

@inproceedings{DBLP:conf/icml/SundararajanTY17,
  author       = {Mukund Sundararajan and
                  Ankur Taly and
                  Qiqi Yan},
  editor       = {Doina Precup and
                  Yee Whye Teh},
  title        = {Axiomatic Attribution for Deep Networks},
  booktitle    = {Proceedings of the 34th International Conference on Machine Learning,
                  {ICML} 2017, Sydney, NSW, Australia, 6-11 August 2017},
  series       = {Proceedings of Machine Learning Research},
  volume       = {70},
  pages        = {3319--3328},
  publisher    = {{PMLR}},
  year         = {2017},
  url          = {http://proceedings.mlr.press/v70/sundararajan17a.html},
  bibsource    = {dblp computer science bibliography, https://dblp.org}
}

@article{DBLP:journals/corr/SmilkovTKVW17,
  author       = {Daniel Smilkov and
                  Nikhil Thorat and
                  Been Kim and
                  Fernanda B. Vi{\'{e}}gas and
                  Martin Wattenberg},
  title        = {SmoothGrad: removing noise by adding noise},
  journal      = {CoRR},
  volume       = {abs/1706.03825},
  year         = {2017},
  url          = {http://arxiv.org/abs/1706.03825},
  eprinttype    = {arXiv},
  eprint       = {1706.03825},
  bibsource    = {dblp computer science bibliography, https://dblp.org}
}

@inproceedings{DBLP:conf/nips/LundbergL17,
  author       = {Scott M. Lundberg and
                  Su{-}In Lee},
  editor       = {Isabelle Guyon and
                  Ulrike von Luxburg and
                  Samy Bengio and
                  Hanna M. Wallach and
                  Rob Fergus and
                  S. V. N. Vishwanathan and
                  Roman Garnett},
  title        = {A Unified Approach to Interpreting Model Predictions},
  booktitle    = {Advances in Neural Information Processing Systems 30: Annual Conference
                  on Neural Information Processing Systems 2017, December 4-9, 2017,
                  Long Beach, CA, {USA}},
  pages        = {4765--4774},
  year         = {2017},
  url_long          = {https://proceedings.neurips.cc/paper/2017/hash/8a20a8621978632d76c43dfd28b67767-Abstract.html},
  bibsource    = {dblp computer science bibliography, https://dblp.org}
}

@inproceedings{DBLP:conf/iccv/FongV17,
  author       = {Ruth C. Fong and
                  Andrea Vedaldi},
  title        = {Interpretable Explanations of Black Boxes by Meaningful Perturbation},
  booktitle    = {{IEEE} International Conference on Computer Vision, {ICCV} 2017, Venice,
                  Italy, October 22-29, 2017},
  pages        = {3449--3457},
  publisher    = {{IEEE} Computer Society},
  year         = {2017},
  url          = {https://doi.org/10.1109/ICCV.2017.371},
  doi          = {10.1109/ICCV.2017.371},
  bibsource    = {dblp computer science bibliography, https://dblp.org}
}

@article{DBLP:journals/jmlr/CovertLL21,
  author       = {Ian Covert and
                  Scott M. Lundberg and
                  Su{-}In Lee},
  title        = {Explaining by Removing: {A} Unified Framework for Model Explanation},
  journal      = {J. Mach. Learn. Res.},
  volume       = {22},
  pages        = {209:1--209:90},
  year         = {2021},
  url          = {https://jmlr.org/papers/v22/20-1316.html},
  bibsource    = {dblp computer science bibliography, https://dblp.org}
}

@inproceedings{DBLP:conf/emnlp/KobayashiKYI21,
  author       = {Goro Kobayashi and
                  Tatsuki Kuribayashi and
                  Sho Yokoi and
                  Kentaro Inui},
  editor       = {Marie{-}Francine Moens and
                  Xuanjing Huang and
                  Lucia Specia and
                  Scott Wen{-}tau Yih},
  title        = {Incorporating Residual and Normalization Layers into Analysis of Masked
                  Language Models},
  booktitle    = {Proceedings of the 2021 Conference on Empirical Methods in Natural
                  Language Processing, {EMNLP} 2021, Virtual Event / Punta Cana, Dominican
                  Republic, 7-11 November, 2021},
  pages        = {4547--4568},
  publisher    = {Association for Computational Linguistics},
  year         = {2021},
  url          = {https://doi.org/10.18653/v1/2021.emnlp-main.373},
  doi          = {10.18653/V1/2021.EMNLP-MAIN.373},
  bibsource    = {dblp computer science bibliography, https://dblp.org}
}

@inproceedings{DBLP:conf/acl/FerrandoGTC23,
  author       = {Javier Ferrando and
                  Gerard I. G{\'{a}}llego and
                  Ioannis Tsiamas and
                  Marta R. Costa{-}juss{\`{a}}},
  editor       = {Anna Rogers and
                  Jordan L. Boyd{-}Graber and
                  Naoaki Okazaki},
  title        = {Explaining How Transformers Use Context to Build Predictions},
  booktitle    = {Proceedings of the 61st Annual Meeting of the Association for Computational
                  Linguistics (Volume 1: Long Papers), {ACL} 2023, Toronto, Canada,
                  July 9-14, 2023},
  pages        = {5486--5513},
  publisher    = {Association for Computational Linguistics},
  year         = {2023},
  url          = {https://doi.org/10.18653/v1/2023.acl-long.301},
  doi          = {10.18653/V1/2023.ACL-LONG.301},
  bibsource    = {dblp computer science bibliography, https://dblp.org}
}

@article{bach2015pixel,
  title={On pixel-wise explanations for non-linear classifier decisions by layer-wise relevance propagation},
  author={Bach, Sebastian and Binder, Alexander and Montavon, Gr{\'e}goire and Klauschen, Frederick and M{\"u}ller, Klaus-Robert and Samek, Wojciech},
  journal={PloS one},
  volume={10},
  number={7},
  pages={e0130140},
  year={2015},
  publisher={Public Library of Science San Francisco, CA USA}
}

@inproceedings{DBLP:conf/icml/AchtibatHDJWLS24,
  author       = {Reduan Achtibat and
                  Sayed Mohammad Vakilzadeh Hatefi and
                  Maximilian Dreyer and
                  Aakriti Jain and
                  Thomas Wiegand and
                  Sebastian Lapuschkin and
                  Wojciech Samek},
  title        = {AttnLRP: Attention-Aware Layer-Wise Relevance Propagation for Transformers},
  booktitle    = {Forty-first International Conference on Machine Learning, {ICML} 2024,
                  Vienna, Austria, July 21-27, 2024},
  publisher    = {OpenReview.net},
  year         = {2024},
  url          = {https://openreview.net/forum?id=emtXYlBrNF},
  bibsource    = {dblp computer science bibliography, https://dblp.org}
}

@misc{arras2025closelookdecompositionbasedxaimethods,
      title={A Close Look at Decomposition-based XAI-Methods for Transformer Language Models}, 
      author={Leila Arras and Bruno Puri and Patrick Kahardipraja and Sebastian Lapuschkin and Wojciech Samek},
      year={2025},
      eprint={2502.15886},
      archivePrefix={arXiv},
      primaryClass={cs.CL},
      url={https://arxiv.org/abs/2502.15886}, 
}

@inproceedings{DBLP:conf/icml/AliSEMMW22,
  author       = {Ameen Ali and
                  Thomas Schnake and
                  Oliver Eberle and
                  Gr{\'{e}}goire Montavon and
                  Klaus{-}Robert M{\"{u}}ller and
                  Lior Wolf},
  editor       = {Kamalika Chaudhuri and
                  Stefanie Jegelka and
                  Le Song and
                  Csaba Szepesv{\'{a}}ri and
                  Gang Niu and
                  Sivan Sabato},
  title        = {{XAI} for Transformers: Better Explanations through Conservative Propagation},
  booktitle    = {International Conference on Machine Learning, {ICML} 2022, 17-23 July
                  2022, Baltimore, Maryland, {USA}},
  series       = {Proceedings of Machine Learning Research},
  volume       = {162},
  pages        = {435--451},
  publisher    = {{PMLR}},
  year         = {2022},
  url          = {https://proceedings.mlr.press/v162/ali22a.html},
  bibsource    = {dblp computer science bibliography, https://dblp.org}
}

@inproceedings{DBLP:conf/emnlp/YinN22,
  author       = {Kayo Yin and
                  Graham Neubig},
  editor       = {Yoav Goldberg and
                  Zornitsa Kozareva and
                  Yue Zhang},
  title        = {Interpreting Language Models with Contrastive Explanations},
  booktitle    = {Proceedings of the 2022 Conference on Empirical Methods in Natural
                  Language Processing, {EMNLP} 2022, Abu Dhabi, United Arab Emirates,
                  December 7-11, 2022},
  pages        = {184--198},
  publisher    = {Association for Computational Linguistics},
  year         = {2022},
  url          = {https://doi.org/10.18653/v1/2022.emnlp-main.14},
  doi          = {10.18653/V1/2022.EMNLP-MAIN.14},
  bibsource    = {dblp computer science bibliography, https://dblp.org}
}

@inproceedings{DBLP:conf/iclr/LiuKR25,
  author       = {Fengyuan Liu and
                  Nikhil Kandpal and
                  Colin Raffel},
  title        = {AttriBoT: {A} Bag of Tricks for Efficiently Approximating Leave-One-Out
                  Context Attribution},
  booktitle    = {The Thirteenth International Conference on Learning Representations,
                  {ICLR} 2025, Singapore, April 24-28, 2025},
  publisher    = {OpenReview.net},
  year         = {2025},
  url          = {https://openreview.net/forum?id=9kJperA2a4},
  bibsource    = {dblp computer science bibliography, https://dblp.org}
}

@inproceedings{DBLP:conf/emnlp/FerrandoV24,
  author       = {Javier Ferrando and
                  Elena Voita},
  editor       = {Yaser Al{-}Onaizan and
                  Mohit Bansal and
                  Yun{-}Nung Chen},
  title        = {Information Flow Routes: Automatically Interpreting Language Models
                  at Scale},
  booktitle    = {Proceedings of the 2024 Conference on Empirical Methods in Natural
                  Language Processing, {EMNLP} 2024, Miami, FL, USA, November 12-16,
                  2024},
  pages        = {17432--17445},
  publisher    = {Association for Computational Linguistics},
  year         = {2024},
  url          = {https://doi.org/10.18653/v1/2024.emnlp-main.965},
  doi          = {10.18653/V1/2024.EMNLP-MAIN.965},
  bibsource    = {dblp computer science bibliography, https://dblp.org}
}

@misc{heimersheim2024useinterpretactivationpatching,
      title={How to use and interpret activation patching}, 
      author={Stefan Heimersheim and Neel Nanda},
      year={2024},
      eprint={2404.15255},
      archivePrefix={arXiv},
      primaryClass={cs.LG},
      url={https://arxiv.org/abs/2404.15255}, 
}

@inproceedings{DBLP:conf/iclr/ZhangN24,
  author       = {Fred Zhang and
                  Neel Nanda},
  title        = {Towards Best Practices of Activation Patching in Language Models:
                  Metrics and Methods},
  booktitle    = {The Twelfth International Conference on Learning Representations,
                  {ICLR} 2024, Vienna, Austria, May 7-11, 2024},
  publisher    = {OpenReview.net},
  year         = {2024},
  url          = {https://openreview.net/forum?id=Hf17y6u9BC},
  bibsource    = {dblp computer science bibliography, https://dblp.org}
}

@article{nanda2023attribution,
  title={Attribution patching: Activation patching at industrial scale},
  author={Nanda, Neel},
  url={https://www.neelnanda.io/mechanistic-interpretability/attribution-patching},
  year={2023}
}

@article{DBLP:journals/corr/abs-2403-00745,
  author       = {J{\'{a}}nos Kram{\'{a}}r and
                  Tom Lieberum and
                  Rohin Shah and
                  Neel Nanda},
  title        = {AtP*: An efficient and scalable method for localizing {LLM} behaviour
                  to components},
  journal      = {CoRR},
  volume       = {abs/2403.00745},
  year         = {2024},
  url          = {https://doi.org/10.48550/arXiv.2403.00745},
  doi          = {10.48550/ARXIV.2403.00745},
  eprinttype    = {arXiv},
  eprint       = {2403.00745},
  bibsource    = {dblp computer science bibliography, https://dblp.org}
}

@inproceedings{DBLP:conf/iclr/WangVCSS23,
  author       = {Kevin Ro Wang and
                  Alexandre Variengien and
                  Arthur Conmy and
                  Buck Shlegeris and
                  Jacob Steinhardt},
  title        = {Interpretability in the Wild: a Circuit for Indirect Object Identification
                  in {GPT-2} Small},
  booktitle    = {The Eleventh International Conference on Learning Representations,
                  {ICLR} 2023, Kigali, Rwanda, May 1-5, 2023},
  publisher    = {OpenReview.net},
  year         = {2023},
  url          = {https://openreview.net/forum?id=NpsVSN6o4ul},
  bibsource    = {dblp computer science bibliography, https://dblp.org}
}

@misc{goldowskydill2023localizingmodelbehaviorpath,
      title={Localizing Model Behavior with Path Patching}, 
      author={Nicholas Goldowsky-Dill and Chris MacLeod and Lucas Sato and Aryaman Arora},
      year={2023},
      eprint={2304.05969},
      archivePrefix={arXiv},
      primaryClass={cs.LG},
      url={https://arxiv.org/abs/2304.05969}, 
}

@inproceedings{DBLP:conf/nips/DunefskyCN24,
  author       = {Jacob Dunefsky and
                  Philippe Chlenski and
                  Neel Nanda},
  editor       = {Amir Globersons and
                  Lester Mackey and
                  Danielle Belgrave and
                  Angela Fan and
                  Ulrich Paquet and
                  Jakub M. Tomczak and
                  Cheng Zhang},
  title        = {Transcoders find interpretable {LLM} feature circuits},
  booktitle    = {Advances in Neural Information Processing Systems 38: Annual Conference
                  on Neural Information Processing Systems 2024, NeurIPS 2024, Vancouver,
                  BC, Canada, December 10 - 15, 2024},
  year         = {2024},
  url_long          = {http://papers.nips.cc/paper\_files/paper/2024/hash/2b8f4db0464cc5b6e9d5e6bea4b9f308-Abstract-Conference.html},
  bibsource    = {dblp computer science bibliography, https://dblp.org}
}

@article{ameisen2025circuit,
  author={Ameisen, Emmanuel and Lindsey, Jack and Pearce, Adam and Gurnee, Wes and Turner, Nicholas L. and Chen, Brian and Citro, Craig and Abrahams, David and Carter, Shan and Hosmer, Basil and Marcus, Jonathan and Sklar, Michael and Templeton, Adly and Bricken, Trenton and McDougall, Callum and Cunningham, Hoagy and Henighan, Thomas and Jermyn, Adam and Jones, Andy and Persic, Andrew and Qi, Zhenyi and Ben Thompson, T. and Zimmerman, Sam and Rivoire, Kelley and Conerly, Thomas and Olah, Chris and Batson, Joshua},
  title={Circuit Tracing: Revealing Computational Graphs in Language Models},
  journal={Transformer Circuits Thread},
  year={2025},
  url={https://transformer-circuits.pub/2025/attribution-graphs/methods.html}
}

@misc{zhao2025verifyingchainofthoughtreasoningcomputational,
      title={Verifying Chain-of-Thought Reasoning via Its Computational Graph}, 
      author={Zheng Zhao and Yeskendir Koishekenov and Xianjun Yang and Naila Murray and Nicola Cancedda},
      year={2025},
      eprint={2510.09312},
      archivePrefix={arXiv},
      primaryClass={cs.CL},
      url={https://arxiv.org/abs/2510.09312}, 
}

@article{lindsey2025biology,
  author={Lindsey, Jack and Gurnee, Wes and Ameisen, Emmanuel and Chen, Brian and Pearce, Adam and Turner, Nicholas L. and Citro, Craig and Abrahams, David and Carter, Shan and Hosmer, Basil and Marcus, Jonathan and Sklar, Michael and Templeton, Adly and Bricken, Trenton and McDougall, Callum and Cunningham, Hoagy and Henighan, Thomas and Jermyn, Adam and Jones, Andy and Persic, Andrew and Qi, Zhenyi and Thompson, T. Ben and Zimmerman, Sam and Rivoire, Kelley and Conerly, Thomas and Olah, Chris and Batson, Joshua},
  title={On the Biology of a Large Language Model},
  journal={Transformer Circuits Thread},
  year={2025},
  url={https://transformer-circuits.pub/2025/attribution-graphs/biology.html}
}

@inproceedings{hanna-etal-2025-circuit,
    title = "Circuit-Tracer: A New Library for Finding Feature Circuits",
    author = "Hanna, Michael  and
      Piotrowski, Mateusz  and
      Lindsey, Jack  and
      Ameisen, Emmanuel",
    editor = "Belinkov, Yonatan  and
      Mueller, Aaron  and
      Kim, Najoung  and
      Mohebbi, Hosein  and
      Chen, Hanjie  and
      Arad, Dana  and
      Sarti, Gabriele",
    booktitle = "Proceedings of the 8th BlackboxNLP Workshop: Analyzing and Interpreting Neural Networks for NLP",
    month = nov,
    year = "2025",
    address = "Suzhou, China",
    publisher = "Association for Computational Linguistics",
    url = "https://aclanthology.org/2025.blackboxnlp-1.14/",
    doi = "10.18653/v1/2025.blackboxnlp-1.14",
    pages = "239--249",
    ISBN = "979-8-89176-346-3"
}

@misc{rosser2025streamscalingmechanisticinterpretability,
      title={Stream: Scaling up Mechanistic Interpretability to Long Context in LLMs via Sparse Attention}, 
      author={J Rosser and José Luis Redondo García and Gustavo Penha and Konstantina Palla and Hugues Bouchard},
      year={2025},
      eprint={2510.19875},
      archivePrefix={arXiv},
      primaryClass={cs.CL},
      url={https://arxiv.org/abs/2510.19875}, 
}

@misc{dai2025graphghosttracingstructureslarge,
      title={GraphGhost: Tracing Structures Behind Large Language Models}, 
      author={Xinnan Dai and Kai Guo and Chung-Hsiang Lo and Shenglai Zeng and Jiayuan Ding and Dongsheng Luo and Subhabrata Mukherjee and Jiliang Tang},
      year={2025},
      eprint={2510.08613},
      archivePrefix={arXiv},
      primaryClass={cs.CL},
      url={https://arxiv.org/abs/2510.08613}, 
}

@misc{yang2025qwen3technicalreport,
      title={Qwen3 Technical Report}, 
      author={An Yang and Anfeng Li and Baosong Yang and Beichen Zhang and Binyuan Hui and Bo Zheng and Bowen Yu and Chang Gao and Chengen Huang and Chenxu Lv and Chujie Zheng and Dayiheng Liu and Fan Zhou and Fei Huang and Feng Hu and Hao Ge and Haoran Wei and Huan Lin and Jialong Tang and Jian Yang and Jianhong Tu and Jianwei Zhang and Jianxin Yang and Jiaxi Yang and Jing Zhou and Jingren Zhou and Junyang Lin and Kai Dang and Keqin Bao and Kexin Yang and Le Yu and Lianghao Deng and Mei Li and Mingfeng Xue and Mingze Li and Pei Zhang and Peng Wang and Qin Zhu and Rui Men and Ruize Gao and Shixuan Liu and Shuang Luo and Tianhao Li and Tianyi Tang and Wenbiao Yin and Xingzhang Ren and Xinyu Wang and Xinyu Zhang and Xuancheng Ren and Yang Fan and Yang Su and Yichang Zhang and Yinger Zhang and Yu Wan and Yuqiong Liu and Zekun Wang and Zeyu Cui and Zhenru Zhang and Zhipeng Zhou and Zihan Qiu},
      year={2025},
      eprint={2505.09388},
      archivePrefix={arXiv},
      primaryClass={cs.CL},
      url={https://arxiv.org/abs/2505.09388}, 
}

@misc{olmo2025olmo3,
      title={Olmo 3}, 
      author={Team Olmo and : and Allyson Ettinger and Amanda Bertsch and Bailey Kuehl and David Graham and David Heineman and Dirk Groeneveld and Faeze Brahman and Finbarr Timbers and Hamish Ivison and Jacob Morrison and Jake Poznanski and Kyle Lo and Luca Soldaini and Matt Jordan and Mayee Chen and Michael Noukhovitch and Nathan Lambert and Pete Walsh and Pradeep Dasigi and Robert Berry and Saumya Malik and Saurabh Shah and Scott Geng and Shane Arora and Shashank Gupta and Taira Anderson and Teng Xiao and Tyler Murray and Tyler Romero and Victoria Graf and Akari Asai and Akshita Bhagia and Alexander Wettig and Alisa Liu and Aman Rangapur and Chloe Anastasiades and Costa Huang and Dustin Schwenk and Harsh Trivedi and Ian Magnusson and Jaron Lochner and Jiacheng Liu and Lester James V. Miranda and Maarten Sap and Malia Morgan and Michael Schmitz and Michal Guerquin and Michael Wilson and Regan Huff and Ronan Le Bras and Rui Xin and Rulin Shao and Sam Skjonsberg and Shannon Zejiang Shen and Shuyue Stella Li and Tucker Wilde and Valentina Pyatkin and Will Merrill and Yapei Chang and Yuling Gu and Zhiyuan Zeng and Ashish Sabharwal and Luke Zettlemoyer and Pang Wei Koh and Ali Farhadi and Noah A. Smith and Hannaneh Hajishirzi},
      year={2025},
      eprint={2512.13961},
      archivePrefix={arXiv},
      primaryClass={cs.CL},
      url={https://arxiv.org/abs/2512.13961}, 
}

@misc{ma2025doverinterventiondrivenautodebugging,
      title={DoVer: Intervention-Driven Auto Debugging for LLM Multi-Agent Systems}, 
      author={Ming Ma and Jue Zhang and Fangkai Yang and Yu Kang and Qingwei Lin and Tianming Yang and Saravan Rajmohan and Dongmei Zhang},
      year={2025},
      eprint={2512.06749},
      archivePrefix={arXiv},
      primaryClass={cs.AI},
      url={https://arxiv.org/abs/2512.06749}, 
}

@misc{nanda2022transformerlens,
    title = {TransformerLens},
    author = {Neel Nanda and Joseph Bloom},
    year = {2022},
    howpublished = {\url{https://github.com/TransformerLensOrg/TransformerLens}},
}

@misc{andrews2025arescalingagentenvironments,
      title={ARE: Scaling Up Agent Environments and Evaluations},
      author={Pierre Andrews and Amine Benhalloum and Gerard Moreno-Torres Bertran and Matteo Bettini and Amar Budhiraja and Ricardo Silveira Cabral and Virginie Do and Romain Froger and Emilien Garreau and Jean-Baptiste Gaya and Hugo Laurençon and Maxime Lecanu and Kunal Malkan and Dheeraj Mekala and Pierre Ménard and Grégoire Mialon and Ulyana Piterbarg and Mikhail Plekhanov and Mathieu Rita and Andrey Rusakov and Thomas Scialom and Vladislav Vorotilov and Mengjue Wang and Ian Yu},
      year={2025},
      eprint={2509.17158},
      archivePrefix={arXiv},
      primaryClass={cs.AI},
      url={https://arxiv.org/abs/2509.17158},
}

@inproceedings{xu2025diagnosing,
  title={Diagnosing failures in large language models’ answers: Integrating error attribution into evaluation framework},
  author={Xu, Zishan and Xie, Shuyi and Lv, Qingsong and Xiao, Shupei and Song, Linlin and Wenjuan, Sui and Lin, Fan},
  booktitle={Findings of the Association for Computational Linguistics: ACL 2025},
  pages={21148--21165},
  year={2025}
}

@misc{ashury2026errormap,
      title={ErrorMap and ErrorAtlas: Charting the Failure Landscape of Large Language Models}, 
      author={Shir Ashury-Tahan and Yifan Mai and Elron Bandel and Michal Shmueli-Scheuer and Leshem Choshen},
      year={2026},
      eprint={2601.15812},
      archivePrefix={arXiv},
      primaryClass={cs.AI},
      url={https://arxiv.org/abs/2601.15812}, 
}

@inproceedings{
hendrycksmath2021,
title={Measuring Mathematical Problem Solving With the {MATH} Dataset},
author={Dan Hendrycks and Collin Burns and Saurav Kadavath and Akul Arora and Steven Basart and Eric Tang and Dawn Song and Jacob Steinhardt},
booktitle={Thirty-fifth Conference on Neural Information Processing Systems Datasets and Benchmarks Track (Round 2)},
year={2021},
url={https://openreview.net/forum?id=7Bywt2mQsCe}
}

@misc{zhou2023instructionfollowingevaluationlargelanguage,
      title={Instruction-Following Evaluation for Large Language Models}, 
      author={Jeffrey Zhou and Tianjian Lu and Swaroop Mishra and Siddhartha Brahma and Sujoy Basu and Yi Luan and Denny Zhou and Le Hou},
      year={2023},
      eprint={2311.07911},
      archivePrefix={arXiv},
      primaryClass={cs.CL},
      url={https://arxiv.org/abs/2311.07911}, 
}

@inproceedings{
xiao2024efficient,
title={Efficient Streaming Language Models with Attention Sinks},
author={Guangxuan Xiao and Yuandong Tian and Beidi Chen and Song Han and Mike Lewis},
booktitle={The Twelfth International Conference on Learning Representations},
year={2024},
url={https://openreview.net/forum?id=NG7sS51zVF}
}

@inproceedings{cancedda2024spectral,
  title={Spectral Filters, Dark Signals, and Attention Sinks},
  author={Cancedda, Nicola},
  booktitle={Proceedings of the 62nd Annual Meeting of the Association for Computational Linguistics (Volume 1: Long Papers)},
  pages={4792--4808},
  year={2024}
}

@misc{openai2025gptoss120bgptoss20bmodel,
      title={gpt-oss-120b \& gpt-oss-20b Model Card}, 
      author={OpenAI},
      year={2025},
      eprint={2508.10925},
      archivePrefix={arXiv},
      primaryClass={cs.CL},
      url={https://arxiv.org/abs/2508.10925}, 
}

@misc{openaigpt5,
      title={Introducing GPT-5}, 
      author={OpenAI},
      year={2025},
      url={https://openai.com/index/introducing-gpt-5/}, 
}

@misc{liu2025deepseek,
      title={DeepSeek-V3.2: Pushing the Frontier of Open Large Language Models}, 
      author={DeepSeek-AI and Liu, Aixin and Mei, Aoxue and Lin, Bangcai and Xue, Bing and Wang, Bingxuan and Xu, Bingzheng and Wu, Bochao and Zhang, Bowei and Lin, Chaofan and Dong, Chen and others},
      year={2025},
      eprint={2512.02556},
      archivePrefix={arXiv},
      primaryClass={cs.CL},
      url={https://arxiv.org/abs/2512.02556}, 
}

@misc{team2025kimi,
      title={Kimi K2: Open Agentic Intelligence}, 
      author={Team, Kimi and Bai, Yifan and Bao, Yiping and Chen, Guanduo and Chen, Jiahao and Chen, Ningxin and Chen, Ruijue and Chen, Yanru and Chen, Yuankun and Chen, Yutian and others},
      year={2025},
      eprint={2507.20534},
      archivePrefix={arXiv},
      primaryClass={cs.LG},
      url={https://arxiv.org/abs/2507.20534}, 
}

@inproceedings{evalplus,
  title = {Is Your Code Generated by Chat{GPT} Really Correct? Rigorous Evaluation of Large Language Models for Code Generation},
  author = {Liu, Jiawei and Xia, Chunqiu Steven and Wang, Yuyao and Zhang, Lingming},
  booktitle = {Thirty-seventh Conference on Neural Information Processing Systems},
  year = {2023},
  url = {https://openreview.net/forum?id=1qvx610Cu7},
}

@inproceedings{evalperf,
  title = {Evaluating Language Models for Efficient Code Generation},
  author = {Liu, Jiawei and Xie, Songrun and Wang, Junhao and Wei, Yuxiang and Ding, Yifeng and Zhang, Lingming},
  booktitle = {First Conference on Language Modeling},
  year = {2024},
  url = {https://openreview.net/forum?id=IBCBMeAhmC},
}

@article{rafailov2023direct,
  title={Direct preference optimization: Your language model is secretly a reward model},
  author={Rafailov, Rafael and Sharma, Archit and Mitchell, Eric and Manning, Christopher D and Ermon, Stefano and Finn, Chelsea},
  journal={Advances in neural information processing systems},
  volume={36},
  pages={53728--53741},
  year={2023}
}

@misc{TIDPO2026,
      title={Token-Importance Guided Direct Preference Optimization}, 
      author={Ning Yang and Hai Lin and Yibo Liu and Baoliang Tian and Guoqing Liu and Haijun Zhang},
      year={2025},
      eprint={2505.19653},
      archivePrefix={arXiv},
      primaryClass={cs.AI},
      url={https://arxiv.org/abs/2505.19653}, 
}
\bibliographystyle{icml2026}

%%%%%%%%%%%%%%%%%%%%%%%%%%%%%%%%%%%%%%%%%%%%%%%%%%%%%%%%%%%%%%%%%%%%%%%%%%%%%%%
%%%%%%%%%%%%%%%%%%%%%%%%%%%%%%%%%%%%%%%%%%%%%%%%%%%%%%%%%%%%%%%%%%%%%%%%%%%%%%%
% APPENDIX
%%%%%%%%%%%%%%%%%%%%%%%%%%%%%%%%%%%%%%%%%%%%%%%%%%%%%%%%%%%%%%%%%%%%%%%%%%%%%%%
%%%%%%%%%%%%%%%%%%%%%%%%%%%%%%%%%%%%%%%%%%%%%%%%%%%%%%%%%%%%%%%%%%%%%%%%%%%%%%%
\newpage
\appendix
\onecolumn

\section{Batch-Packed Multi-Target Backpropagation for Attribution Graph Construction}
\label{app:batch_packed_backprop}

This appendix presents an efficient procedure for constructing attribution-graph edges under
Layer-wise Relevance Propagation (LRP).
The method leverages recent results showing that LRP-style relevance propagation in transformer
models can be implemented via a \emph{modified gradient$\times$input} formulation, enabling direct
use of automatic differentiation frameworks such as PyTorch
\cite{DBLP:conf/icml/AchtibatHDJWLS24,arras2025closelookdecompositionbasedxaimethods}.
Building on this equivalence, we further exploit the batch dimension to propagate relevance from
multiple attribution targets simultaneously, substantially reducing computational cost.

\paragraph{Why gradients represent relevance.}
LRP aims to decompose a scalar attribution objective, such as a contrastive logit difference
$\Delta \ell$ as in this work, into contributions from intermediate activations.
For non-linear architectures like transformers, recent work has shown that LRP rules can be
implemented by \emph{patching the backward pass} with customized local propagation rules.
Crucially, these patched backward passes are equivalent to computing a modified
gradient$\times$input quantity, allowing relevance to be obtained using standard automatic
differentiation
\cite{DBLP:conf/icml/AchtibatHDJWLS24,arras2025closelookdecompositionbasedxaimethods}.

Under this formulation, the relevance of a hidden state
$\mathbf{h}_l^{(i)} \in \mathbb{R}^d$ is approximated by
\begin{equation}
\mathbf{R}_l^{(i)}
\;\approx\;
\mathbf{h}_l^{(i)} \odot
\frac{\partial \Delta \ell}{\partial \mathbf{h}_l^{(i)}},
\end{equation}
where the gradient indicates how sensitive the attribution objective is to the hidden state.
Thus, gradients act as the carriers of relevance, and no explicit relevance propagation rules need
to be implemented beyond the modified backward pass.

\paragraph{Problem setup.}
We seek to construct an \emph{attribution graph} whose nodes correspond to hidden-state relevances
and whose edges describe how relevance propagates between layers.
Fix a source layer $s$ and a target layer $t$ with $s < t$.
Let
\begin{equation}
\mathbf{H}^{(s)} \in \mathbb{R}^{{\color{orange}{1}} \times n \times d}
\end{equation}
denote the source-layer hidden states for a sequence of length $n$, and let
\begin{equation}
\mathbf{H}^{(t)} = F_{s \rightarrow t}\!\left(\mathbf{H}^{(s)}\right)
\in \mathbb{R}^{1 \times n \times d}
\end{equation}
denote the corresponding target-layer hidden states, where
$F_{s \rightarrow t}$ is the composition of transformer submodules between layers $s$ and $t$.
%(including the appropriate MHA and MLP subcomponents).

A single forward and backward pass from $\Delta \ell$ yields gradients
\begin{equation}
\mathbf{g}_l^{(i)} = \frac{\partial \Delta \ell}{\partial \mathbf{h}_l^{(i)}}
\end{equation}
at every hidden state.
These gradients determine node relevances via gradient$\times$input and are stored for later use.
At this stage, we know \emph{which} hidden states are relevant, but not \emph{how} relevance flows
between hidden states across layers.

\paragraph{Edge attribution via gradient$\times$input.}
To construct attribution-graph edges, we propagate relevance from a target-layer hidden state
$\mathbf{h}_t^{(j)}$ to all source-layer hidden states $\{\mathbf{h}_s^{(i)}\}_{i=1}^n$.
We treat the stored gradient
$\mathbf{g}_t^{(j)} = \partial \Delta \ell / \partial \mathbf{h}_t^{(j)}$
as a fixed relevance signal.
Because automatic differentiation requires a scalar objective, we define the attribution target
as the inner product
\begin{equation}
\langle \mathbf{h}_t^{(j)}, \mathbf{g}_t^{(j)} \rangle,
\end{equation}
which simply selects and weights the hidden dimensions that are relevant for the final decision.
Backpropagating this scalar through $F_{s \rightarrow t}$ yields gradients with respect to
$\mathbf{H}^{(s)}$.

Relevance propagated from target token $j$ to source token $i$ is then computed using the same
gradient$\times$input rule:
\begin{equation}
A_{j,i}
\;\approx\;
\sum_{k=1}^{d}
\mathbf{h}_s^{(i,k)}
\frac{\partial \langle \mathbf{h}_t^{(j)}, \mathbf{g}_t^{(j)} \rangle}
{\partial \mathbf{h}_s^{(i,k)}}.
\end{equation}
This quantity defines a directed edge from node $(s,i)$ to node $(t,j)$ in the attribution graph.

\paragraph{Batch-packed multi-target backpropagation.}
Naïvely, computing $A_{j,i}$ for all $j \in \{1,\dots,n\}$ would require $O(n)$ separate backward
passes, one per target token.
To avoid this cost, we exploit the batch dimension to process multiple attribution targets
simultaneously, following recent work on attribution graph construction with sparse features in transcoders~\cite{hanna-etal-2025-circuit}.

Let $\mathcal{J} = \{j_1,\dots,j_B\}$ be a set of target indices.
We construct a batched source tensor
\begin{equation}
\widetilde{\mathbf{H}}^{(s)} \in \mathbb{R}^{{\color{orange}{B}} \times n \times d},
\qquad
\widetilde{\mathbf{H}}^{(s)}_{b,:,:}
=
\mathbf{H}^{(s)}_{1,:,:},
\end{equation}
so that each batch element contains an identical copy of the source-layer hidden states.
Forward propagation yields
\begin{equation}
\widetilde{\mathbf{H}}^{(t)} =
F_{s \rightarrow t}\!\left(\widetilde{\mathbf{H}}^{(s)}\right)
\in \mathbb{R}^{B \times n \times d}.
\end{equation}

We then define a batched output-gradient tensor
$\widetilde{\mathbf{G}}^{(t)} \in \mathbb{R}^{B \times n \times d}$ such that batch element $b$
injects relevance only at its corresponding target index $j_b$:
\begin{equation}
\widetilde{\mathbf{G}}^{(t)}_{b, j_b, :} = \mathbf{g}_t^{(j_b)},
\qquad
\widetilde{\mathbf{G}}^{(t)}_{b, j, :} = \mathbf{0} \;\; \text{for } j \neq j_b.
\end{equation}
A single automatic differentiation call computes the corresponding gradients with respect to
$\widetilde{\mathbf{H}}^{(s)}$, yielding $B$ independent relevance propagations in parallel.
Applying gradient$\times$input and writing each batch result into the appropriate row produces
$B$ rows of the token-to-token interaction matrix.

Thus, by batching attribution targets, we reduce the cost of constructing attribution-graph edges
from $O(n)$ backward passes to $O(\lceil n/B \rceil)$ passes, while remaining fully compatible
with LRP-style relevance propagation through standard automatic differentiation.

\section{Computational Efficiency of Batch-Packed Backpropagation}\label{app:efficiency}

This appendix provides empirical efficiency measurements for the batch-packed multi-target backpropagation method introduced in~\autoref{app:batch_packed_backprop}. All experiments are conducted on a single NVIDIA A100 GPU using Qwen3-0.6B unless otherwise stated.

\paragraph{Scaling with batch size.}
\autoref{tab:efficiency_batch} reports latency and peak GPU memory across different batch sizes $B$ and input token lengths. Each cell shows ``latency (seconds) / memory (MB)''.

\begin{table}[htbp]
\centering
\caption{Latency and memory of batch-packed backpropagation across batch sizes (Qwen3-0.6B).}
\label{tab:efficiency_batch}
\small
\begin{tabular}{ccccc}
\toprule
\textbf{Token length} & $B{=}1$ & $B{=}4$ & $B{=}8$ & $B{=}16$ \\
\midrule
63   & 0.26\,/\,3.5   & 0.08\,/\,14.0  & 0.05\,/\,27.9   & 0.02\,/\,55.8   \\
252  & 1.26\,/\,13.7  & 0.25\,/\,53.7  & 0.14\,/\,107.1  & 0.09\,/\,215.3  \\
754  & 2.63\,/\,42.4  & 1.02\,/\,164.6 & 0.77\,/\,327.0  & 0.63\,/\,643.6  \\
1508 & 6.23\,/\,89.2  & 3.08\,/\,333.1 & 2.67\,/\,648.9  & 3.45\,/\,1290.0 \\
\bottomrule
\end{tabular}
\end{table}

Batching yields significant speedups across diverse input lengths (although the gain diminishes for longer sequences), with memory scaling linearly as expected.

\paragraph{Scaling with model size.}
\autoref{tab:efficiency_model} reports latency and memory across different Qwen3 model sizes at a fixed batch size of $B{=}8$.

\begin{table}[htbp]
\centering
\caption{Latency and memory of batch-packed backpropagation across model sizes ($B{=}8$).}
\label{tab:efficiency_model}
\small
\begin{tabular}{ccccc}
\toprule
\textbf{Token length} & \textbf{Qwen3-0.6B} & \textbf{Qwen3-1.7B} & \textbf{Qwen3-4B} & \textbf{Qwen3-8B} \\
\midrule
63   & 0.04\,/\,27.9   & 0.04\,/\,47.7   & 0.04\,/\,75.4   & 0.04\,/\,92.8   \\
252  & 0.13\,/\,107.1  & 0.13\,/\,189.8  & 0.21\,/\,294.3  & 0.32\,/\,371.1  \\
754  & 0.57\,/\,321.6  & 1.07\,/\,570.6  & 1.86\,/\,882.4  & 3.01\,/\,1117.6 \\
1508 & 2.41\,/\,647.5  & 4.41\,/\,1147.0 & 9.99\,/\,1769.2 & 13.33\,/\,2228.1 \\
\bottomrule
\end{tabular}
\end{table}

Both latency and memory scale roughly proportionally to model size, indicating that the method remains efficient and scalable for larger models.

\section{Attribution Graph Pruning and Connected Subgraph Extraction}
\label{app:graph_pruning}

This appendix describes the pruning strategy used to construct a sparse and interpretable
\emph{attribution graph} from the dense token-to-token interaction matrices produced by
batch-packed multi-target backpropagation
(\autoref{app:batch_packed_backprop}).
The objective of pruning is to remove low-relevance edges while preserving the dominant
relevance flows between hidden states across layers.

\paragraph{Input representation.}
For each ordered pair of layers $(s,t)$ with $s < t$, we assume access to:
\begin{itemize}
    \item a token-to-token interaction matrix
    \begin{equation}
        \mathbf{A}^{(s \rightarrow t)} \in \mathbb{R}^{n \times n},
    \end{equation}
    where $\mathbf{A}^{(s \rightarrow t)}_{j,i}$ denotes the relevance propagated from
    source-layer hidden state $\mathbf{h}_s^{(i)}$ to target-layer hidden state
    $\mathbf{h}_t^{(j)}$, as defined by the gradient$\times$input formulation in~\autoref{app:batch_packed_backprop};
    \item source-layer token relevances
    \begin{equation}
        R_i^{(s)} \;\equiv\; \sum_{k=1}^d \mathbf{R}_s^{(i,k)};
    \end{equation}
    \item target-layer token relevances
    \begin{equation}
        R_j^{(t)} \;\equiv\; \sum_{k=1}^d \mathbf{R}_t^{(j,k)}.
    \end{equation}
\end{itemize}
Each nonzero entry of $\mathbf{A}^{(s \rightarrow t)}$ corresponds to a candidate directed edge
from node $(s,i)$ to node $(t,j)$ in the attribution graph.

\paragraph{Graph construction.}
We construct a directed graph
$\mathcal{G} = (V, E)$,
where each node is indexed by a layer--token pair $(l,i)$.
For each layer transition $(s,t)$, we:
\begin{enumerate}
    \item add all source nodes $(s,i)$ such that $|R_i^{(s)}| > 0$;
    \item add all target nodes $(t,j)$ such that $|R_j^{(t)}| > 0$;
    \item add directed edges $(s,i) \rightarrow (t,j)$ for selected entries
    $\mathbf{A}^{(s \rightarrow t)}_{j,i}$ after pruning, with edge weight
    \begin{equation}
        w_{(s,i)\rightarrow(t,j)} = \mathbf{A}^{(s \rightarrow t)}_{j,i}.
    \end{equation}
\end{enumerate}

\paragraph{Pruning objectives.}
The interaction matrices $\mathbf{A}^{(s \rightarrow t)}$ are generally dense, making direct
graph construction impractical.
We therefore prune edges using magnitude-based criteria applied to
$|\mathbf{A}^{(s \rightarrow t)}|$.
Pruning is performed \emph{independently for each layer pair} to preserve local attribution
structure and avoid conflating relevance scales across layers.

Let
\begin{equation}
E_{\mathrm{nz}}^{(s \rightarrow t)}
=
\{(j,i) \mid |\mathbf{A}^{(s \rightarrow t)}_{j,i}| > 0\}
\end{equation}
denote the set of nonzero candidate edges, and let
\begin{equation}
M^{(s \rightarrow t)}
=
\sum_{(j,i)\in E_{\mathrm{nz}}^{(s \rightarrow t)}}
\left| \mathbf{A}^{(s \rightarrow t)}_{j,i} \right|
\end{equation}
denote the total absolute relevance mass for that layer transition.

\paragraph{Pruning modes.}
We support two pruning strategies:

\begin{itemize}
\item \textit{Global threshold pruning.}
In this mode, we retain all edges whose absolute relevance exceeds a
fixed threshold $\tau > 0$:
\begin{equation}
E^{(s \rightarrow t)}_{\mathrm{keep}}
=
\left\{
(j,i) \;\middle|\;
\left| \mathbf{A}^{(s \rightarrow t)}_{j,i} \right| > \tau
\right\}.
\end{equation}
This mode enforces a uniform relevance scale across all layer transitions and is most appropriate
when attribution magnitudes are comparable across layers.

\item \textit{Per-layer cumulative mass pruning.}
In this mode, pruning is performed adaptively per
layer transition by preserving a fixed fraction of the total relevance mass.
Let
\[
\{a_1 \ge a_2 \ge \dots \ge a_K\}
\]
denote the sorted absolute values of all nonzero entries of
$\mathbf{A}^{(s \rightarrow t)}$, where
$K = |E_{\mathrm{nz}}^{(s \rightarrow t)}|$.
Given a percentile parameter $p \in (0,1]$, we select the smallest index $k^\ast$ such that
\begin{equation}
\sum_{k=1}^{k^\ast} a_k
\;\ge\;
p \cdot \sum_{k=1}^{K} a_k.
\end{equation}
The resulting dynamic threshold is
\begin{equation}
\tau_{s\rightarrow t}^{(p)} = a_{k^\ast},
\end{equation}
and we retain all edges satisfying
\begin{equation}
\left| \mathbf{A}^{(s \rightarrow t)}_{j,i} \right|
\;\ge\;
\tau_{s\rightarrow t}^{(p)}.
\end{equation}
This strategy guarantees that a fixed proportion of relevance mass is preserved for each layer
transition, even when absolute relevance scales differ substantially across layers.

\end{itemize}

\paragraph{Connected attribution subgraph extraction.}
After pruning, the attribution graph may be disconnected.
Since our attribution objective is the contrastive logit difference at the final token position,
we retain only the subgraph that contributes to this prediction.
Specifically, we select the target node corresponding to the hidden state at the last layer and
last token position, i.e., $(L-1,\, n)$, and extract the induced subgraph consisting of all nodes that can reach the target node via directed paths.
All nodes and edges not connected to this target node are discarded.

\paragraph{Resulting graph properties.}
The final attribution graph $\mathcal{G}_{\mathrm{conn}}$ is sparse, directed, and weighted.
Each node $(l,i)$ is annotated with its layer index, token position, and scalar node relevance
$R_i^{(l)}$, while each directed edge encodes signed relevance flow between hidden states across
layers.
By construction, every node and edge in $\mathcal{G}_{\mathrm{conn}}$ lies on at least one
relevance path terminating at the final prediction, enabling faithful path tracing, subgraph
analysis, and visualization of dominant attribution circuits.

\section{More Details on Failure Case Collection}\label{app:failure_case_collection}

%As described in the main text, we disabled the models' thinking mode for IFEval, MATH, and EvalPlus to reduce confounding effects of self-reflection. This appendix reports a controlled study verifying that attribution patterns are robust to this design choice.

Here we provide additional details on failure case collection. For IFEval, MATH, and EvalPlus, we disable the models’ thinking mode to simplify reasoning traces and enable clearer error attribution. Prior work has shown that reflective reasoning can introduce non-monotonic trajectories that complicate analysis~\cite{xu2025diagnosing, ashury2026errormap}. In contrast, for the GAIA2 benchmark, we retain the default configuration with thinking mode enabled, using the official Meta AI ARE system~\cite{andrews2025arescalingagentenvironments}, as GAIA2 involves complex multi-turn tasks where thinking mode supports long-horizon coherence.

To verify that our conclusions are robust to this design choice, we conduct a small controlled study. Specifically, we compare attribution distributions and failure mode classifications with and without thinking mode on all 21 clean IFEval cases using Qwen3-0.6B. Of these, 16 cases yield identical annotations. Among the 5 remaining cases, 1 results in unbounded generation (and is excluded), while the other 4 exhibit shifts in the target–contrast pairs due to altered reasoning trajectories, reflecting changes in model behavior rather than instability in our method. \autoref{tab:thinking_mode} reports the distribution of failure patterns in the two settings. One then observes that the overall attribution patterns are largely preserved: the majority of cases remain consistent, and URT remains the dominant failure pattern in both settings. This confirms that our attribution framework is robust to the thinking mode setting.

\begin{table}[htbp]
\centering
\caption{Failure pattern distribution on IFEval with and without thinking mode (Qwen3-0.6B).}
\label{tab:thinking_mode}
\small
\begin{tabular}{lcccc}
\toprule
\textbf{Setting} & \textbf{\# Cases} & \textbf{URT} & \textbf{OIT} & \textbf{URT+OIT} \\
\midrule
w/o thinking & 20 & 70.0\%  & 5.0\%   & 25.0\%  \\
w/ thinking  & 19 & 58.0\%  & 10.5\%  & 31.5\%  \\
\bottomrule
\end{tabular}
\end{table}

\section{Contrast Token Pair Identification}\label{app:token_pair_identification}
This appendix consolidates the procedures for (i) identifying the first error token (target token) in a model‑generated completion, and (ii) selecting an appropriate contrast token for contrastive attribution. The details on target token agreement statistics and the discussions on the impact of our existing contrast token selection strategy are also presented. 

\subsection{Target Token Identification}
For each failure case, we identify the first output token that causes the reasoning to deviate from the correct trajectory. This token is referred to as the target token. 
We adopt a semi‑automated annotation procedure to identify the target token. Specifically, we prompt multiple powerful LLMs including GPT-5.2-20251211~\cite{openaigpt5}, GPT-OSS-120B~\cite{openai2025gptoss120bgptoss20bmodel}, DeepSeek-V3.2~\cite{liu2025deepseek} and Kimi-K2-Thinking~\cite{team2025kimi}, to independently predict the most likely target token for each failure case. The token receiving the majority of votes is selected as the automated result. Human annotators then verify this prediction and correct it when necessary, balancing efficiency with precision. The prompt template used in the above semi‑automated annotation procedure is unified across all LLMs to ensure deterministic and reproducible identification of the target token. The full template is shown below (using GAIA2 as an example).

\begin{lstlisting}[language=YAML,basicstyle=\footnotesize\ttfamily]
version: v1 
system: |-
  - You are an expert in localizing error tokens in model\-generated text completions. 
  - You will be provided with a prompt and a model\-generated completion, and your task is to localize and output the FIRST token in the completion that is erroneous or causes the error. 
  - As the prompt and completion are taken from a particular dataset, refer to Section "Dataset Description" for a brief overview of the dataset. 
  - Refer to Section "On the Input Format" for details on how the prompt and completion will be presented to you. 
  - Follow the "General Guidance \(Dataset Agnostic\)" section for general instructions on how to approach the task of error token localization. 
  - Refer to the "Dataset Specific Instructions" section for any additional instructions that are specific to this dataset. 
  - Finally, adhere to the "On the Output Format" section to ensure your response is structured correctly. 
  - Note that the dataset specific sections \(i.e., "Dataset Description" and "Dataset Specific Instructions"\) can be missing for some datasets. In such cases, simply skip those sections and proceed with the rest of the instructions. 
  
  # Dataset Description 
    Gaia2 is a benchmark dataset for evaluating AI agents inside simulated, dynamic environments provided by the Meta Agents Research Environments \(ARE\). 
    It contains time\-evolving scenarios with apps/tools \(e.g., messaging, calendar, filesystem, shopping, city, contacts\) and multi\-agent interactions. 
    Agents must plan, search, and act via tool/API calls under temporal constraints, dynamic events, and noise. 
    The dataset evaluates dimensions such as Execution, Search, Adaptability, Time, Ambiguity, and multi\-agent collaboration. 
    For details, see the dataset card and ARE documentation on Gaia2. 
  # On the Input Format 
  - The input will be provided in the following JSON format: 
    { 
    "prompt": "<The prompt text here>", 
    "completion": "<The model\-generated completion text here>", 
    "indexed_completion": "<The model\-generated completion text with each token indexed, i.e., `token_1[index_1]token_2[index_2]`"
    "ground\_truth": "\<The ground truth answer text here\>" \(This field may be absent in some cases\) 
    } 
  - An example of the input format is as follows: 
    {
      "prompt": "<|im_start|>user
      What is the capital of France?<|im_end|>
      <|im_start|>assistant
      ",
      "completion": "<think> okay, let me see. The capital of France is Lyon. </think>
      \Lyon.",
      "indexed_completion": "<think>[0] okay,[1] let[2] me[3] see.[4] The[5] capital[6] of[7] France[8] is[9] Lyon.[10] </think>[11]
      
      [12] Lyon[13] .[14]",
      "ground_truth": "Paris, the capital of France."
    }
  - The "indexed_completion" field provides a tokenized version of the completion, where each token is followed by its index in square brackets. This will help you localize the position of tokens in the completion. 
  - Both the "prompt" and "completion" fields may contain special tokens, e.g., "<|im_start|>", "<|im_end|>", "<think>", "</think>", etc. These tokens are part of the model's output format, specifically,
    - "<|im_start|>" and "<|im_end|>" denote the start and end of a message in a multi-turn conversation, e.g., "<|im_start|>user
    xxx <|im_end|>" indicates the start of a user message, while "<|im_start|>assistant
    xxx <|im_end|>" indicates the start of an assistant message.
    - "<think>" and "</think>" denote the start and end of the model's internal reasoning process.
  - More on the internal thought process of the model: 
  - The internal thought process of the model is represented by the reasoning tokens between the "<think>" and "</think>" tokens. 
  - This part reflects the model's reasoning steps before arriving at the final answer. 
  - The final answer to the prompt is the text that comes AFTER the "</think>" token. 
  # General Guidance (Dataset Agnostic) 
  - Refer to the following steps for localizing an erroneous token in the model-generated completion: 
    - Step 1: Carefully read dataset description and the provided prompt to understand the context and requirements. 
    - Step 2: If "completion" and "indexed_completion" fields contains reasoning tokens (i.e., tokens between "<think>" and "</think>"), seperate the reasoning part from the final answer part. 
    - Step 3: Examine the final answer part of the completion first to identify any errors. If no errors are found in this part, skip the rest of the steps and output "<CORRECT>" as the final answer. 
    - Step 4: No matther whether the final answer part contains errors, proceed to examine the reasoning part for potential errors. 
    - Step 5: Once an erroneous token is identified, use the "indexed_completion" field to find its index. 
  - Addiational strict requirements for localizing the erroneous token:
    - Always report the FIRST erroneous token in the "completion" or "indexed_completion" field.
    - The localized erroneous token MUST NOT be the special tokens used for formatting (e.g., "<|im_start|>", "<|im_end|>", "<think>", "</think>").
    - NEVER treat a token is erroneous just because it is a subword or punctuation mark, as the indexed token str does NOT have to be a single word; it can be a subword or punctuation mark as per the tokenization used by the model.
  - On the difference between examining the final answer part and the reasoning part:
    - For the final answer part (i.e., the part AFTER "</think>"):
      - ALWAYS examine whether the tokens in this part fulfill the format/style requirements specified in the prompt. If you identify a token in this part that violates the format/style requirements, you SHOULD consider it as a potential erroneous token.
      - ALSO examine the factual correctness and logical consistency of the tokens in this part. If you identify a token in this part that is factually incorrect or logically inconsistent with the prompt, you SHOULD consider it as a potential erroneous token.
    - For the reasoning part (i.e., the part BETWEEN "<think>" and "</think>"):
      - NEVER examine whether the reasoning tokens have fulfilled the format/style requirements specified in the prompt, as the reasoning tokens between "<think>" and "</think>" are often in a free-form text format. For example, if the prompt requires the model to "respond in a poem format", you MUST NOT consider a token in the reasoning part as erroneous just because the reasoning tokens are not in a poem format. 
      - Focus on the factual correctness and logical consistency of the reasoning tokens. If you identify a token in this part that is factually incorrect or logically inconsistent with the prompt, you SHOULD consider it as a potential erroneous token.

  # Dataset Specific Instructions 
  - This dataset uses the ARE simulation where agent messages may trigger tool/API calls and receive environment feedback. 
  - Messages labeled as tool outputs (e.g., segments marked as `role=tool-response` with content being tool results) represent environment feedback and MUST NOT be selected as erroneous tokens. Only tokens authored by the assistant are eligible. 
  - Your goal is to localize the FIRST token in the assistant completion that commits the agent to a course of action that makes the user request unachievable in the scenario context. 
  - "Unachievable" includes, but is not limited to: 
  - Invoking a wrong or unavailable tool/app (e.g., writing to filesystem via `create__fs` when exploration to locate a folder is required first). 
  - Assuming nonexistent resources or paths (e.g., hallucinated file/folder IDs or directories). 
  - Attempting UI interactions that the environment does not support (e.g., instructing to "open a user interface" when no such UI exists or is permitted). 
  - Skipping required discovery steps (e.g., failing to locate a folder before listing its contents) and immediately issuing a write/irreversible action. 
  - When such an action spans multiple indexed tokens, localize the very first token that begins the erroneous action or argument (e.g., the first token that starts the wrong tool name or the hallucinated path string). 
  - Tie-break rules: 
  - If both planning text and an ensuing call appear, choose the first token that commits execution (e.g., the token that starts the tool name or the imperative "call"), not earlier speculative text. 
  - If multiple independent errors appear, always select the earliest by index. 
  - IMPORTANT: Ignore input format issues for tools (e.g., minor JSON/schema/quoting/whitespace mistakes) unless they change semantics. Such surface-form errors should not be selected as the FIRST error if the underlying chosen action is already incorrect. 
  - Likewise, ignore any content in `role=tool-response` segments; they are environment outputs, not assistant errors. 
  - If no erroneous behavior is found and the answer satisfies the prompt and scenario constraints, output "<CORRECT>". 
  # On the Output Format 
  - The output MUST be in the following JSON format: 
    { 
    "error_token": "<The localized erroneous token>", 
    "token_index": <The index of the erroneous token in the completion>, 
    "explanation": "<A brief explanation of why this token is considered erroneous. Less than 50 words.>" 
    } 
  - An example of the output format is as follows: 
    { 
    "error_token": "Lyon", 
    "token_index": 10, 
    "explanation": "The token 'Lyon' is erroneous because the correct capital of France is Paris, not Lyon. This indicates a factual error in the model's completion." 
    } 
  - Ensure that the "token_index" corresponds to the index provided in the "indexed_completion" field. 
  - The "explanation" should be concise yet informative, providing enough context to justify the localization of the error token.
\end{lstlisting}

\subsection{Target Token Agreement Statistics}
\label{app:target_token_agreement}

Here we report quantitative evidence for the reliability of our target token identification procedure described in the main text. We employ a two-stage pipeline: (1)~four independent LLMs each propose the target token, and (2)~human annotators validate cases with majority ($\geq$2) agreement.

\paragraph{Inter-agreement among LLM Proposers.}
We use four independent validators: DeepSeek-V3.2~\cite{liu2025deepseek}, Kimi-K2-Thinking~\cite{team2025kimi}, GPT-5.2~\cite{openaigpt5}, and GPT-OSS-120b~\cite{openai2025gptoss120bgptoss20bmodel}. \autoref{tab:target_agreement} reports the fraction of samples reaching $\geq$2-validator consensus within $\pm\delta$ tokens, alongside the chance baseline for each benchmark. It shows that agreement rates are far above the chance baseline across all benchmarks, indicating strong consistency among independent LLM proposers.

\begin{table}[htbp]
\centering
\caption{LLM inter-validator agreement on target token identification. ``Random'' denotes the chance baseline; ``Exact ($\pm 0$)'' and ``Relaxed ($\pm 3$)'' denote the fraction of samples with $\geq$2-validator consensus within the respective token window.}
\label{tab:target_agreement}
\small
\begin{tabular}{lccc}
\toprule
\textbf{Benchmark} & \textbf{Random} & \textbf{Exact ($\pm 0$)} & \textbf{Relaxed ($\pm 3$)} \\
\midrule
EvalPlus & 7.1\% & 57.4\% & 86.7\% \\
IFEval   & 7.7\% & 74.3\% & 89.2\% \\
MATH     & 1.2\% & 56.0\% & 88.0\% \\
\bottomrule
\end{tabular}
\end{table}

\paragraph{Human validation.}
We further conduct human validation on the LLM majority-agreed cases. \autoref{tab:human_validation} reports the fraction of cases approved by human annotators under $\pm 3$-token matching. The high approval rates confirm that LLM consensus aligns well with human judgment, validating the reliability of our semi-automated target token identification pipeline.

\begin{table}[htbp]
\centering
\caption{Human approval rates for LLM majority-agreed target token identifications.}
\label{tab:human_validation}
\small
\begin{tabular}{lcc}
\toprule
\textbf{Benchmark} & \textbf{Exact-agreed} & \textbf{Relaxed-agreed} \\
\midrule
EvalPlus & 80.0\% & 97.3\% \\
IFEval   & 96.4\% & 90.1\% \\
MATH     & 78.6\% & 86.4\% \\
\bottomrule
\end{tabular}
\end{table}

\subsection{Contrast Token Identification}
To analyze the model’s preference mechanism, we pair the target token with a contrast token, a plausible alternative that would have led to a more correct continuation. The selection follows the following steps:
\begin{enumerate}
    \item Assuming the target token is the model’s top-1 output token at the failure position (i.e., the token with the highest logit).
    \item Examining the model’s top-10 predicted tokens at that position. If any of these tokens, when substituted, would plausibly lead to a more correct continuation, we select the highest-ranked such token as the contrast token.
    \item If no suitable contrast token is found in the top-10 predictions, we query a stronger sibling model (e.g., Qwen3-1.7B for Qwen3-0.6B, or Qwen3-8B for Qwen3-4B) and inspect its top-10 predictions at the same position. If a valid contrast token is found, we again select the highest-ranked one.
    \item If neither model yields a suitable candidate, we manually infer a semantically appropriate contrast token. In cases where multiple plausible candidates exist, we choose the one with the highest rank in the original model’s output logits.
\end{enumerate}

\subsection{Impact of Contrast Token Selection and Plausible Alternatives}
\label{app:alternative_contrast_token_pairs}

The strict requirement on contrast tokens ensures high-confidence contrastive pairs for attribution, but it also constitutes the primary source of case filtering. \autoref{tab:filtering_breakdown} summarizes the filtering breakdown across different processing stages. We observe that the dominant filtering factor is the inability to identify a valid contrast token under the strict recovery criterion, rather than ambiguity in target-token identification or failure to meet the logit threshold.
%To further verify that the retained subset is representative, we nearly tripled the data (from 71 to 196 cases) and compared attribution patterns between the original and extended batches; chi-squared tests of homogeneity confirm no significant distributional shift (all $p > 0.05$), with the dominant attribution outcome (M-IA) and failure pattern (URT) preserved across all benchmarks.

\begin{table}[h]
\centering
\caption{Filtering breakdown by stage (\% of original failure cases excluded at each stage).}
\label{tab:filtering_breakdown}
\small
\begin{tabular}{lcccc}
\toprule
\textbf{Stage} & \textbf{IFEval} & \textbf{MATH} & \textbf{EvalPlus} & \textbf{GAIA2} \\
\midrule
Unbounded generation & 17.3\% & 6.7\% & 0\% & 12\% \\
Unable to identify target token & 0\% & 6.7\% & 43.3\% & 5\% \\
\textbf{Unable to identify contrast token} & \textbf{67.3\%} & \textbf{37.7\%} & \textbf{36.8\%} & \textbf{68\%} \\
Logit-diff threshold & 1.9\% & 13.3\% & 0\% & 0\% \\
\bottomrule
\end{tabular}
\end{table}

To investigate whether we could relax such strict requirement on contrast token selection, we constructed three additional contrast-token pairs per case on a random subset of 12 failure cases; on average, \textbf{7.51 out of 10} top positively contributing tokens overlap with those from the baseline pair, indicating that qualitative conclusions are stable across plausible alternatives.

This suggests that the strict recovery requirement could, in principle, be relaxed to admit alternative contrast tokens, even when the model cannot fully recover due to limited capability. However, we adopt the stricter criterion to ensure clean, high-confidence contrastive pairs, and leave the incorporation of more flexible alternatives to future work.

\section{Inter-Annotator Agreement Statistics on Attribution Analysis}\label{app:annotator_agreement}

We also report inter-annotator agreement statistics for the two key annotation stages in our analysis: (1)~the classification of attribution outcomes (M-IA, NC-IA+M-AG, NC-IA+AG) and (2)~the classification of failure patterns (URT, OIT, URT+OIT). Two annotators independently labeled all cases following a standardized annotation protocol consisting of three phases: a \emph{pre-calibration} phase to assess initial agreement and identify sources of disagreement, a \emph{calibration discussion} to align category definitions and resolve ambiguities, and a \emph{post-calibration} phase to measure agreement after alignment.

\autoref{tab:iaa_outcomes} reports agreement rates and Wilson 95\% confidence intervals for the classification of attribution outcomes, while \autoref{tab:iaa_patterns} reports agreement rates for the classification of failure patterns among cases categorized as M-IA or NC-IA+M-AG. One can see that after calibration, agreement improved substantially to 88.2\% for attribution outcomes and 83.3\% for failure patterns. All results reported in the main text use post-calibration annotations.

\begin{table}[htbp]
\centering
\caption{Inter-annotator agreement on attribution outcome classification.}
\label{tab:iaa_outcomes}
\small
\begin{tabular}{llccc}
\toprule
\textbf{Phase} & \textbf{Dataset} & $N$ & \textbf{Agreement} & \textbf{Wilson 95\% CI} \\
\midrule
Pre-calibration  & IFEval    & 21 & 71.4\% & [50.0\%, 86.2\%] \\
Pre-calibration  & MATH      & 16 & 50.0\% & [28.0\%, 72.0\%] \\
Pre-calibration  & EvalPlus  & 19 & 47.4\% & [27.3\%, 68.3\%] \\
\textbf{Post-calibration} & \textbf{EvalPlus} & \textbf{34} & \textbf{88.2\%} & \textbf{[73.4\%, 95.3\%]} \\
\bottomrule
\end{tabular}
\end{table}

\begin{table}[htbp]
\centering
\caption{Inter-annotator agreement on failure pattern classification.}
\label{tab:iaa_patterns}
\small
\begin{tabular}{llccc}
\toprule
\textbf{Phase} & \textbf{Dataset} & $N$ & \textbf{Agreement} & \textbf{Wilson 95\% CI} \\
\midrule
Pre-calibration  & IFEval    & 18 & 38.9\% & [20.3\%, 61.4\%] \\
Pre-calibration  & MATH      & 5  & 40.0\% & [11.8\%, 76.9\%] \\
Pre-calibration  & EvalPlus  & 12 & 58.3\% & [32.0\%, 80.7\%] \\
\textbf{Post-calibration} & \textbf{EvalPlus} & \textbf{24} & \textbf{83.3\%} & \textbf{[64.1\%, 93.3\%]} \\
\bottomrule
\end{tabular}
\end{table}

\section{Comparison with Alternative Attribution Methods}\label{app:method_comparison}

This appendix provides a direct comparison between AttnLRP (the attribution method adopted in this work) and two representative baselines: an alternative LRP-based method (CP-LRP~\cite{DBLP:conf/icml/AliSEMMW22}) and a standard gradient-based method (Gradient~\cite{DBLP:journals/corr/SimonyanVZ13}). Our goal is not to introduce a new attribution method, but to explore if AttnLRP provides faithful attributions suitable for failure diagnosis and that the diagnostic framework is compatible with alternative faithful attribution methods.

\paragraph{Perturbation-based evaluation.}
We evaluate all three methods using a perturbation-based protocol on 60 randomly selected failure cases across all four datasets (IFEval, EvalPlus, MATH, and GAIA2). For each case, we rank input tokens by attribution score and measure how many top-ranked tokens must be masked to flip the model's top-1 prediction away from the error token. \autoref{tab:perturbation} reports the average number of tokens required and the overall fix rate.

\begin{table}[htbp]
\centering
\caption{Perturbation-based evaluation of attribution methods across 60 failure cases.}
\label{tab:perturbation}
\small
\begin{tabular}{lcc}
\toprule
\textbf{Method} & \textbf{Avg.\ tokens to fix} & \textbf{Fix rate} \\
\midrule
AttnLRP  & 1.7 & 100\%  \\
CP-LRP   & 2.0 & 100\%  \\
Gradient & 3.7 & 96.7\% \\
\bottomrule
\end{tabular}
\end{table}

Both LRP-based methods achieve a 100\% fix rate, whereas the gradient baseline still fails in some cases. Between the two LRP methods, the difference is marginal (1.7 vs.\ 2.0 tokens), consistent with the original AttnLRP paper where the faithfulness gap between AttnLRP and CP-LRP is also small.

\paragraph{Attribution sharpness.}
Beyond perturbation accuracy, we measure how concentrated each method's attribution is on the most relevant tokens. We compute two metrics: (1)~the \emph{concentration ratio}, defined as the fraction of total attribution mass captured by the top-10 content tokens (excluding special tokens such as \texttt{<|im\_start|>} and \texttt{<think>}), and (2)~the \emph{Gini coefficient}, measuring the overall inequality of the attribution distribution ($1.0$ = all mass in one token, $0.0$ = uniform).

\begin{table}[htbp]
\centering
\caption{Concentration ratio (top-10 content tokens).}
\label{tab:concentration}
\small
\begin{tabular}{lccc}
\toprule
\textbf{Method} & \textbf{Mean} & \textbf{Median} & \textbf{Std} \\
\midrule
AttnLRP  & 0.5176 & 0.4889 & 0.1611 \\
CP-LRP   & 0.5111 & 0.4738 & 0.1916 \\
Gradient & 0.4428 & 0.4198 & 0.1669 \\
\bottomrule
\end{tabular}
\end{table}

\begin{table}[htbp]
\centering
\caption{Gini coefficient of attribution distributions.}
\label{tab:gini}
\small
\begin{tabular}{lccc}
\toprule
\textbf{Method} & \textbf{Mean} & \textbf{Median} & \textbf{Std} \\
\midrule
AttnLRP  & 0.7326 & 0.7398 & 0.0664 \\
CP-LRP   & 0.7061 & 0.7262 & 0.1046 \\
Gradient & 0.6817 & 0.6798 & 0.0629 \\
\bottomrule
\end{tabular}
\end{table}

AttnLRP achieves the highest mean and median on both metrics, indicating a marginally sharper signal. It also exhibits the lowest Gini standard deviation (0.066 vs.\ 0.105 for CP-LRP), suggesting more stable attribution sharpness across cases. Both LRP methods substantially outperform Gradient, whose more diffuse attributions are consistent with its lower perturbation efficiency.

\paragraph{Qualitative comparison.}
%To illustrate the qualitative differences, consider an IFEval case in which the model is asked to produce a Japan itinerary in Shakespearean style without using commas. The model erroneously inserts a comma after ``halls''. We analyze input attributions for the decision between the target token ``\texttt{,}'' and the contrast token ``\texttt{ and}''. \emph{Gradient} assigns the strongest signals to largely irrelevant content/style tokens (e.g., ``Japan'', ``Shakespearean''). In contrast, \emph{AttnLRP} and \emph{CP-LRP} assign higher importance to constraint-related tokens (e.g., ``not allowed'', ``commas''), with AttnLRP exhibiting a sharper and more localized focus. This suggests that gradient-based attributions are less faithful in this setting, whereas LRP-based methods better capture the governing constraint.
To illustrate the qualitative differences, consider an IFEval case in which the model is asked to produce a Japan itinerary in Shakespearean style without using commas: 

\emph{``I am planning a trip to Japan, and I would like thee to write an itinerary for my journey in a Shakespearean style. You are not allowed to use any commas in your response.''}

The model produces an incorrect output by inserting a comma after “halls”: 

\emph{``In the land of thy golden halls,''}

We analyze input attributions for the decision between the target token “,” and an alternative contrast token “ and”. Gradient assigns the strongest signals to largely irrelevant content/style tokens (e.g., “Japan”, “Shakespearean”). In contrast, AttnLRP and CP-LRP assign higher importance to constraint-related tokens (e.g., “not allowed”, “commas”), with AttnLRP exhibiting a sharper and more localized focus. This suggests that gradient-based attributions are less faithful in this setting, whereas LRP-based methods better capture the governing constraint, with AttnLRP providing a more precise signal for debugging and improvement.

Overall, we acknowledge that AttnLRP and CP-LRP perform comparably in this diagnostic setting, and either could serve as the underlying attribution method. Our choice of AttnLRP is motivated by its slightly sharper signal and stronger performance reported in the original paper. Our contribution lies in the diagnostic framework rather than the specific attribution method, and the framework is compatible with any faithful attribution method.

\section{Sample Expanded Attribution Graphs} \label{app:AG}

\begin{figure}[htbp]
    \centering
    \includegraphics[width=0.92\columnwidth]{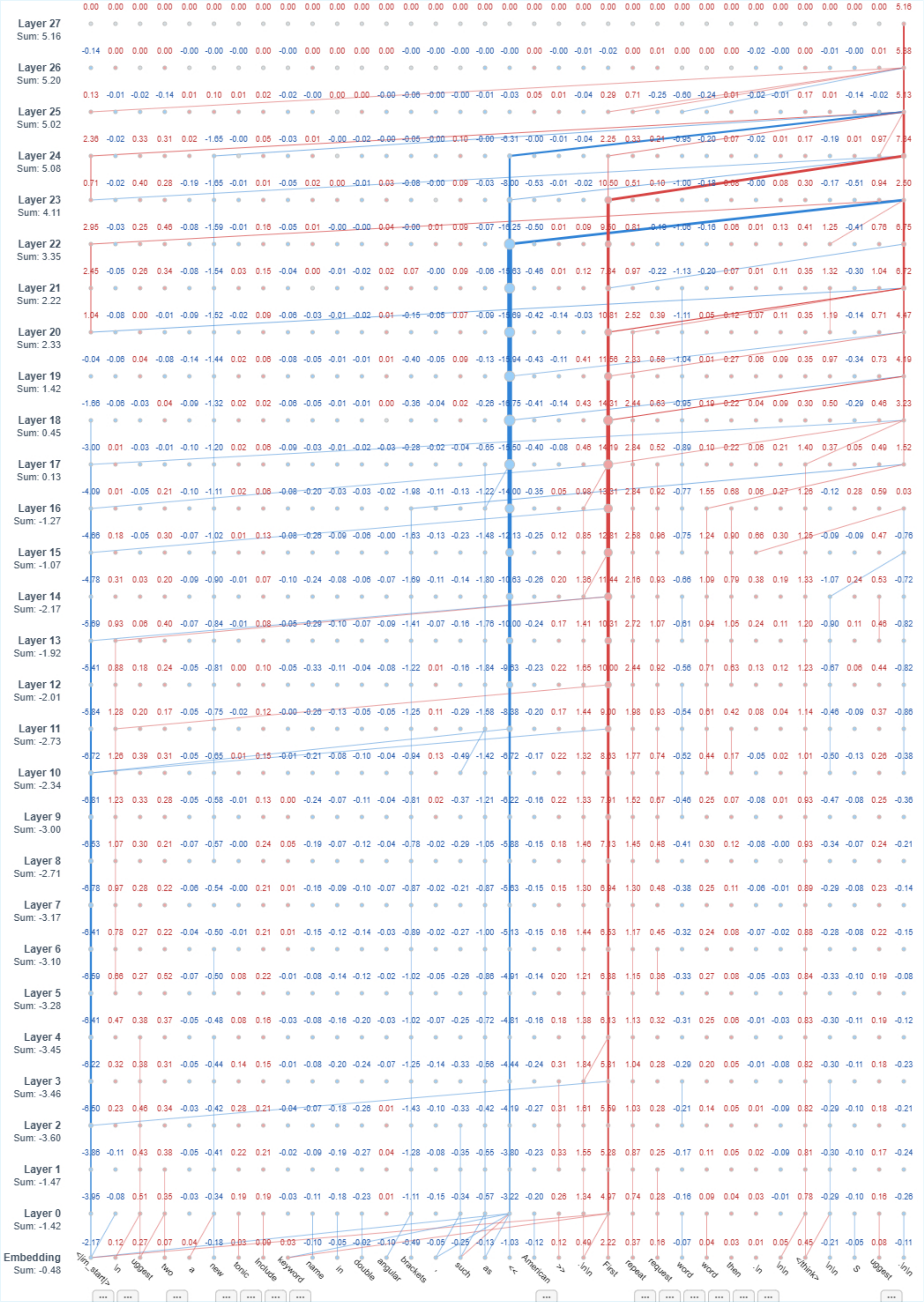}
    \caption{Expanded attribution graph for case in \autoref{fig:AG}.}
    \label{fig:fullag}
\end{figure}

For clarity of exposition, the attribution graph shown in the main text is a reduced representation that includes only the dominant layers, nodes, and relevance pathways. Minor layers and low‑magnitude edges are omitted to avoid visual clutter.
The \autoref{fig:fullag} provides an expanded version of the graph that preserves a larger portion of the model’s internal relevance propagation. While the full graph still prunes extremely low‑contribution edges for readability, it offers a more exhaustive view of the attribution structure used in our analysis.

\begin{figure}[htbp]
    \centering
    \includegraphics[width=0.85\columnwidth]{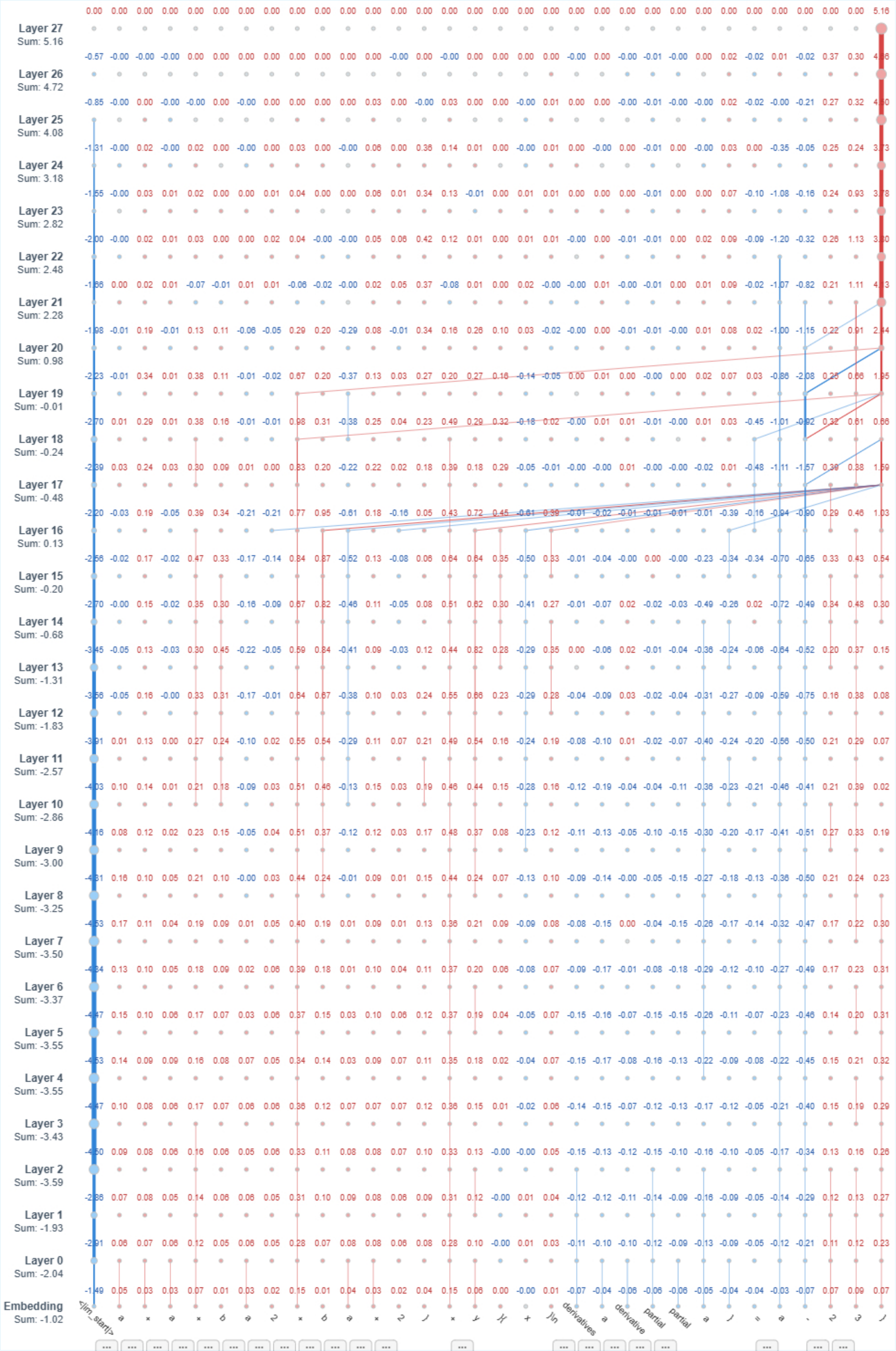}
    \caption{Expanded attribution graph for an example case from MATH, with biases embedded in internal representations.}
    \label{fig:bos_0}
\end{figure}

In addition to the representative cases discussed above, we also observe another class of attribution graph patterns at MATH dataset, as shown in \autoref{fig:bos_0}. It further illuminates the model’s internal decision dynamics. Specifically, the full attribution graph for this example reveals that the model’s preference for the target token is driven overwhelmingly by internal biases, most notably the BOS token, while receiving only minimal contribution from the reasoning chain present in the prompt. The relevance mass rapidly concentrates along a small set of bias‑dominated pathways, bypassing the majority of intermediate reasoning tokens. This structure indicates that the model’s erroneous decision does not arise from misinterpreting the logical steps in the prompt, but rather from an inherent predisposition encoded in deep layers and positional biases, which suppress the semantic evidence that should have been propagated along the reasoning chain.

% Another case from GAIA2 is shown in \autoref{fig:bos_1}. This example drawn from a long-horizon GAIA2 task, further demonstrates how the model’s internal decision dynamics can be dominated by structural biases rather than contextual reasoning. Specifically, even after pruning low‑relevance nodes and edges for visibility, the expanded attribution graph reveals that the preference for the target token is shaped primarily by the BOS‑driven pathways and a small set of deeply embedded biases, while the extensive reasoning chain in the prompt contributes only marginally to the final decision. The relevance mass converges rapidly onto these bias‑dominated routes across layers, indicating that the model arrives at its erroneous output not by misinterpreting the underlying reasoning process, but by relying on internal predispositions that override the semantic evidence provided by the task context.

\section{Statistical Analysis of Attribution Graph Structure}\label{app:ag_statistics}

This appendix presents an exploratory statistical analysis of the attribution graphs constructed for the failure cases studied in the main text. We examine the layer-wise relevance structure of the final (prediction) token, decompose its attribution into distinct source components, and investigate whether these structural features give rise to clearly separable failure categories. The analysis covers all 57 clean failure traces (21 from IFEval, 19 from EvalPlus, 17 from MATH) for input-level relevance profiling, of which 23 are categorized as NC-IA and undergo detailed attribution graph analysis (Sections~K.1--K.4). Cross-model comparisons use Qwen3-0.6B, 1.7B, 4B, and 8B on the same failure cases (Section~K.5). As we show below, while certain aggregate-level regularities emerge robustly (e.g., layer-wise functional specialization and consistent critical-layer locations), the data do not exhibit cleanly separable clusters along any single structural dimension, highlighting the challenge of automated failure categorization from attribution graph structure alone.

\subsection{Relevance Profile Analysis}

For each failure case, we extract the scalar relevance $R_i^{(l)}$ of the last token (i.e., the token position at which the next-token prediction is made) across all $L$ transformer layers, yielding a \emph{relevance profile} vector of dimension $L+1$ (including the embedding layer). For Qwen3-0.6B, the raw relevance at the embedding layer averages $0.66$ and grows monotonically to $18.85$ at the final layer (L27), confirming that the contrastive logit difference is progressively amplified across the transformer stack. To enable comparison across cases with different logit-difference magnitudes, we normalize each profile by subtracting the embedding-layer value and dividing by the absolute value of the final-layer relevance:
\begin{equation}
\hat{R}^{(l)} = \frac{R^{(l)} - R^{(-1)}}{|R^{(L-1)}|},
\end{equation}
where $R^{(-1)}$ and $R^{(L-1)}$ denote the embedding-layer and final-layer relevance, respectively. The resulting normalized profiles reveal how relevance accumulates across the transformer stack for different failure cases.

\begin{figure}[htbp]
    \centering
    \includegraphics[width=\columnwidth]{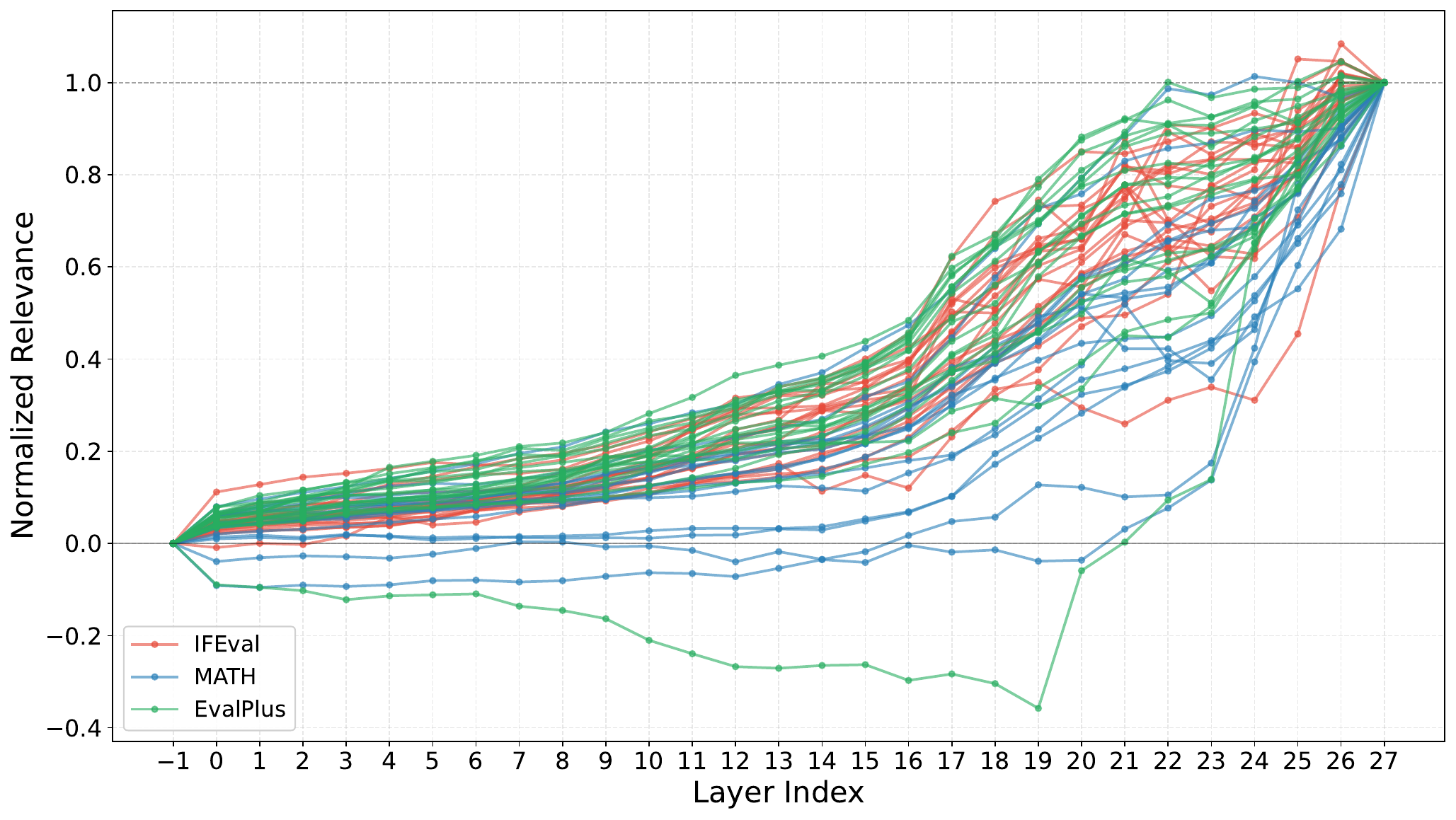}
    \caption{Normalized relevance profiles of the prediction token across all failure cases, colored by dataset. Each curve shows how relevance accumulates from the embedding layer (L$-1$) to the final layer (L27).}
    \label{fig:normalized_profiles}
\end{figure}

\autoref{fig:normalized_profiles} visualizes the normalized relevance profiles for all failure cases across the three benchmarks (IFEval, MATH, and EvalPlus). While all profiles share a monotonically increasing trend by construction, distinct trajectory shapes emerge: some cases exhibit rapid early growth followed by saturation, whereas others show a delayed, late-layer surge. To investigate whether these shape differences correspond to meaningful failure categories, we perform an exploratory clustering analysis.

\begin{figure}[htbp]
    \centering
    \includegraphics[width=\columnwidth]{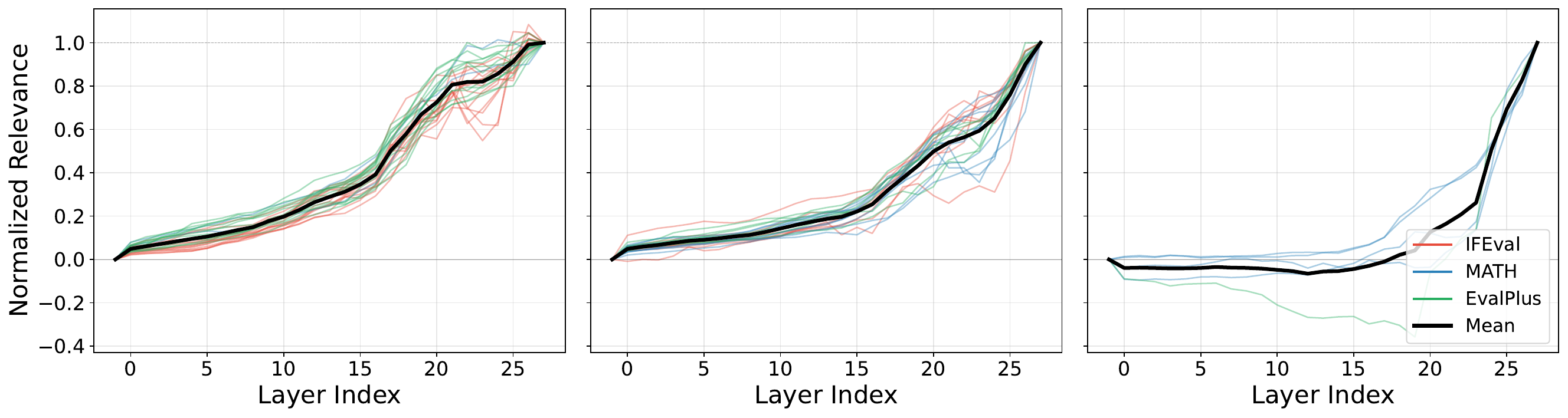}
    \caption{Clustered relevance profiles ($k{=}3$). Each panel shows individual traces (thin lines) and the cluster mean (thick line).}
    \label{fig:profile_clusters}
\end{figure}

\begin{figure}[htbp]
    \centering
    \includegraphics[width=0.75\columnwidth]{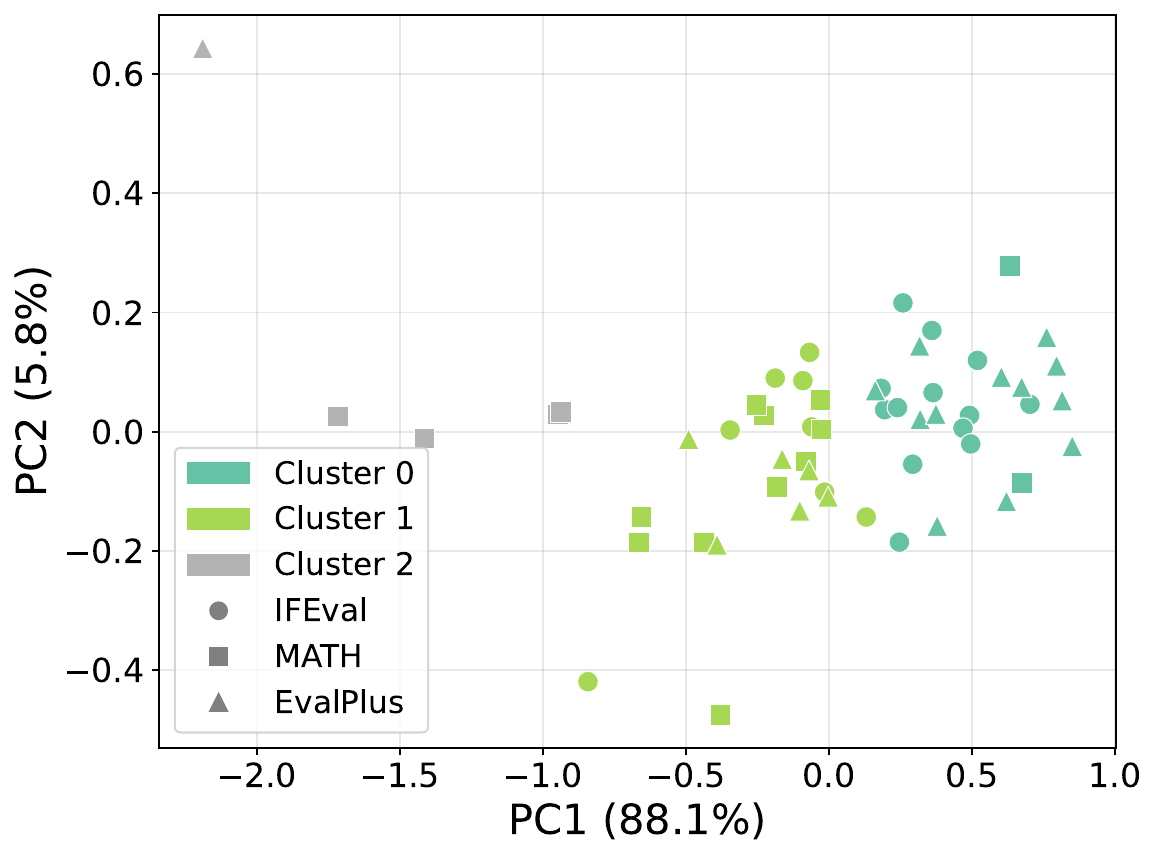}
    \caption{PCA 2D projection of the 29-dimensional normalized relevance profiles, colored by $k$-means cluster assignment.}
    \label{fig:profile_pca}
\end{figure}

\paragraph{Exploratory clustering of relevance profiles.}
Since relevance profile analysis and attribution decomposition require attribution graph construction, which is not performed for cases already explained by input attribution (M-IA), we restrict the following analyses to the 23 failure traces categorized as NC-IA (with or without subsequent attribution graph explanation). We apply $k$-means clustering to the normalized relevance profile vectors (each 29-dimensional), selecting $k{=}3$ via the elbow curvature and silhouette score. The resulting clusters (\autoref{fig:profile_clusters}) suggest three broad accumulation patterns: (i)~gradual near-linear growth (Cluster~0, $n{=}13$), (ii)~early plateau followed by late-layer acceleration (Cluster~1, $n{=}3$), and (iii)~rapid early growth with gentle late-layer progression (Cluster~2, $n{=}7$). A PCA projection shows reasonable separation in the first two principal components (\autoref{fig:profile_pca}), though as we show in the next subsection, this separation does not extend to other structural dimensions. We also caution that Cluster~1 contains only 3 traces, limiting the statistical reliability of any conclusions drawn from it; the clustering should therefore be interpreted as descriptive rather than as evidence of discrete failure types.

\subsection{Attribution Decomposition}\label{app:ag_decomposition}

To understand the sources of relevance at the prediction token, we decompose the attribution at each layer into three components using the edge structure of the attribution graph:
\begin{itemize}[leftmargin=*]
    \item \textbf{Self Bias (SB):} The residual relevance at the prediction token not accounted for by any incoming edges, i.e., the difference between the node relevance and the total incoming edge weight. This captures the model's internal bias at that position.
    \item \textbf{BOS Contribution (BOS):} The total edge weight from the beginning-of-sequence token (position 0) to the prediction token. This captures the influence of the attention sink~\cite{xiao2024efficient}.
    \item \textbf{Other-Token Contribution (OC):} The total edge weight from all other tokens (excluding BOS and self-loops) to the prediction token. This captures the influence of context tokens.
\end{itemize}

Formally, for the prediction token at position $n$ and layer $l$:
\begin{equation}
R_n^{(l)} = \underbrace{\text{SB}^{(l)}}_{\text{self bias}} + \underbrace{\text{BOS}^{(l)}}_{\text{BOS contribution}} + \underbrace{\text{OC}^{(l)}}_{\text{other contribution}},
\end{equation}
where $\text{BOS}^{(l)} = \sum_{e \in \mathcal{E}_{\text{BOS} \to n}^{(l)}} w_e$, $\text{OC}^{(l)} = \sum_{e \in \mathcal{E}_{\text{other} \to n}^{(l)}} w_e$, and $\text{SB}^{(l)} = R_n^{(l)} - \text{BOS}^{(l)} - \text{OC}^{(l)}$.

To obtain scale-invariant composition features, we normalize each component by the mean absolute total relevance $\overline{|R|} = \frac{1}{L+1}\sum_{l} |R_n^{(l)}|$ and define the composition proportions:
\begin{equation}
\text{SB fraction} = \frac{\overline{\text{SB}}}{\overline{\text{SB}} + \overline{\text{OC}}}, \quad
\text{BOS fraction} = \frac{|\overline{\text{BOS}}|}{\overline{\text{SB}} + \overline{\text{OC}}},
\end{equation}
where $\overline{(\cdot)}$ denotes the layer-wise mean of the normalized values. The SB fraction measures the relative dominance of self bias over context integration, while the BOS fraction measures the relative strength of the attention-sink effect.

\begin{figure}[htbp]
    \centering
    \includegraphics[width=0.75\columnwidth]{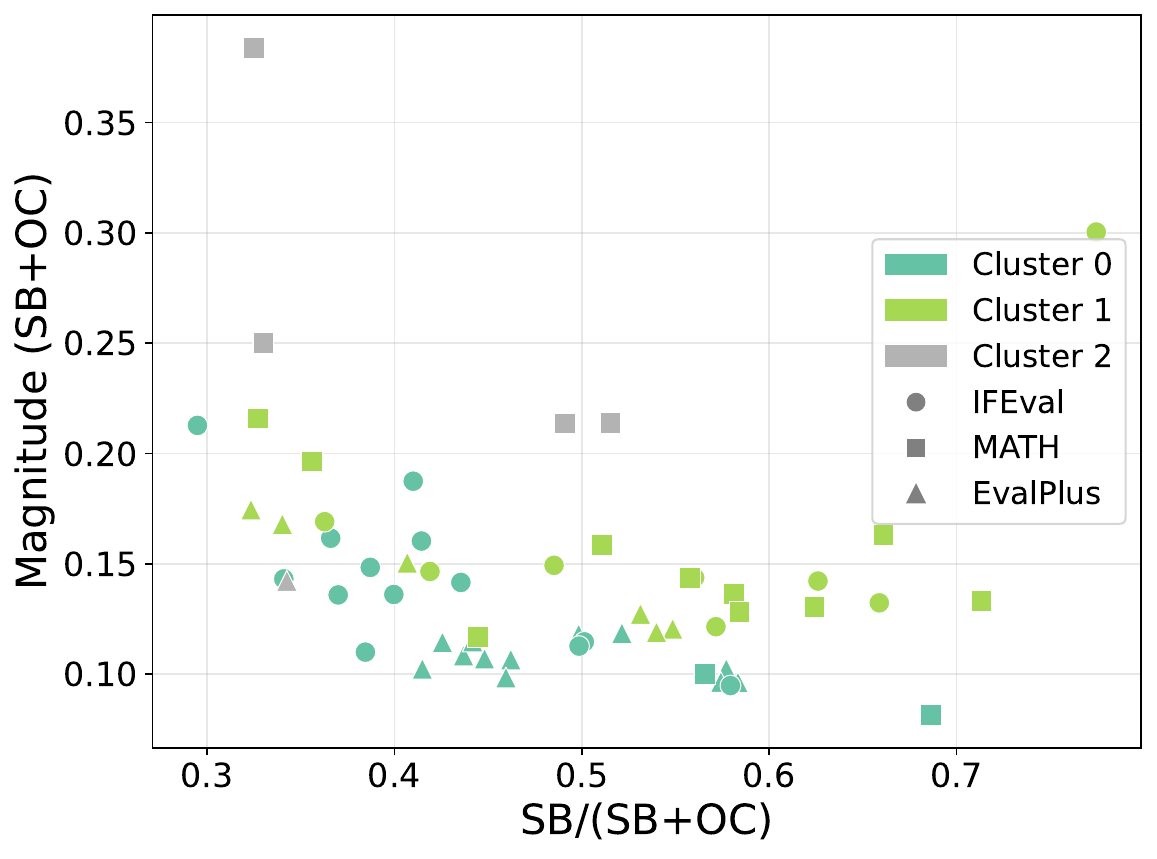}
    \caption{Composition space: SB fraction vs.\ total magnitude (SB+OC), colored by relevance-profile cluster. Marker shapes indicate datasets.}
    \label{fig:composition_scatter}
\end{figure}

\autoref{fig:composition_scatter} shows the composition space as a scatter plot of SB fraction versus total magnitude, colored by profile cluster. The per-cluster mean composition statistics are shown in \autoref{tab:cluster_composition}.

\begin{table}[htbp]
\centering
\caption{Per-cluster mean composition statistics (mean $\pm$ std).}
\label{tab:cluster_composition}
\small
\begin{tabular}{lccc}
\toprule
\textbf{Cluster} & \textbf{SB/(SB+OC)} & \textbf{$|$BOS$|$/(SB+OC)} & \textbf{Magnitude} \\
\midrule
Cluster 0 ($n{=}13$) & $0.516 \pm 0.119$ & $0.239 \pm 0.112$ & $0.164 \pm 0.036$ \\
Cluster 1 ($n{=}3$)  & $0.332 \pm 0.009$ & $0.339 \pm 0.055$ & $0.259 \pm 0.121$ \\
Cluster 2 ($n{=}7$)  & $0.528 \pm 0.108$ & $0.258 \pm 0.098$ & $0.104 \pm 0.017$ \\
\bottomrule
\end{tabular}
\end{table}

\paragraph{Profile shape and composition are orthogonal.}
A key finding is that the relevance-profile clusters do \emph{not} align with the composition structure. The intra-cluster variance is $2.61\times$ and $8.80\times$ larger than the inter-cluster variance for SB fraction and BOS fraction, respectively, as shown in \autoref{tab:variance_ratio}. Moreover, direct $k$-means clustering on the composition space yields near-zero agreement with profile-based clusters (Adjusted Rand Index $= -0.004$). This orthogonality demonstrates that the attribution graph structure is inherently multi-dimensional: \emph{how} relevance accumulates across layers (profile shape) is largely independent of \emph{where} the relevance originates (SB vs.\ BOS vs.\ context). As a consequence, no single clustering scheme can adequately capture the structural diversity of failure cases, underscoring the difficulty of building automated failure taxonomies from attribution graphs at the current scale.

\begin{table}[htbp]
\centering
\caption{Intra-cluster vs.\ inter-cluster variance ratio for composition features.}
\label{tab:variance_ratio}
\small
\begin{tabular}{lccc}
\toprule
\textbf{Feature} & \textbf{Intra-$\sigma^2$} & \textbf{Inter-$\sigma^2$} & \textbf{Ratio} \\
\midrule
SB/(SB+OC)       & 0.0105 & 0.0040 & 2.61$\times$ \\
$|$BOS$|$/(SB+OC) & 0.0093 & 0.0011 & 8.80$\times$ \\
Magnitude (SB+OC) & 0.0020 & 0.0022 & 0.90$\times$ \\
\bottomrule
\end{tabular}
\end{table}

\subsection{Layer-Wise Functional Specialization}

\begin{figure}[htbp]
    \centering
    \includegraphics[width=\columnwidth]{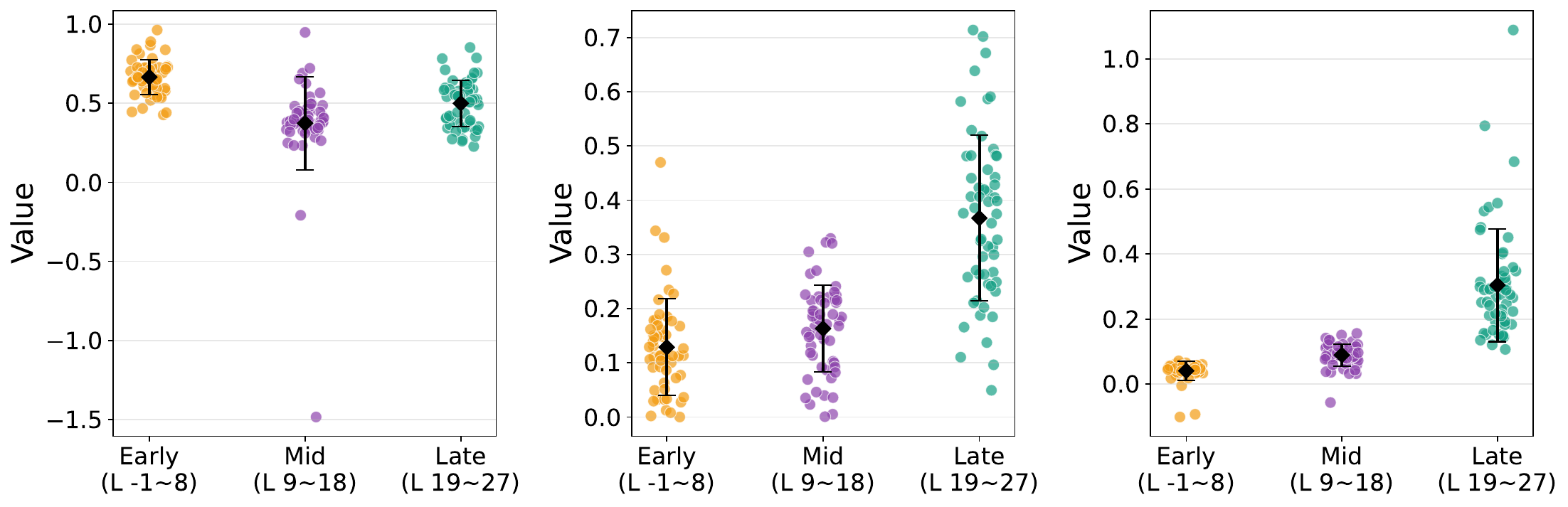}
    \caption{Distribution of composition features by layer segment (Early/Mid/Late). Diamonds denote segment means $\pm$ 1 standard deviation; dots show individual traces.}
    \label{fig:layer_granularity}
\end{figure}

In contrast to the clustering analyses above, which reveal high structural heterogeneity across individual traces, aggregating composition features by \emph{layer segment} exposes a robust and consistent pattern. We partition the transformer layers into three segments: \emph{Early} (layers $-1$ to $8$), \emph{Mid} (layers $9$ to $18$), and \emph{Late} (layers $19$ to $27$).\footnote{Layer indices are specific to Qwen3-0.6B (28 transformer layers, indexed L0--L27). For models with different depths, the segments are adjusted proportionally.} We then compute the mean composition features within each segment (\autoref{fig:layer_granularity}).

Unlike the case-level clustering, this aggregate analysis yields clear and consistent findings across all 23 traces:
\begin{itemize}[leftmargin=*]
    \item \textbf{Self bias dominates early layers.} The SB fraction is highest in early layers (mean $\approx 0.73$) and decreases to $\approx 0.49$--$0.51$ in mid and late layers, indicating that early-layer computations are primarily driven by position-specific biases rather than contextual information.
    \item \textbf{Context integration peaks in mid layers.} The other-token contribution (OC fraction $\approx 0.51$) is highest in mid layers, indicating that the model's integration of contextual information is most active in the middle of the transformer stack. This finding is consistent with prior observations that mid-layer attention heads are responsible for semantic composition~\cite{DBLP:conf/emnlp/KobayashiKYI21}.
    \item \textbf{BOS influence grows in later layers.} The BOS fraction increases from early ($\approx 0.15$) to late layers ($\approx 0.27$--$0.36$), consistent with the attention-sink phenomenon becoming more pronounced in deeper layers.
\end{itemize}

This three-phase functional specialization (\emph{early bias establishment, mid-layer context integration, late-layer bias amplification}) is the most robust structural finding in our analysis: it holds across all traces regardless of profile cluster, dataset, or decomposition pattern. It also provides a mechanistic explanation for the main-text findings: failures amenable to input-level attribution (M-IA) correspond to cases where mid-layer context integration goes awry, while failures requiring attribution-graph analysis (NC-IA+M-AG) tend to involve late-layer bias accumulation that overrides contextual signals.

\subsection{Critical Layer Identification}

\begin{figure}[htbp]
    \centering
    \includegraphics[width=\columnwidth]{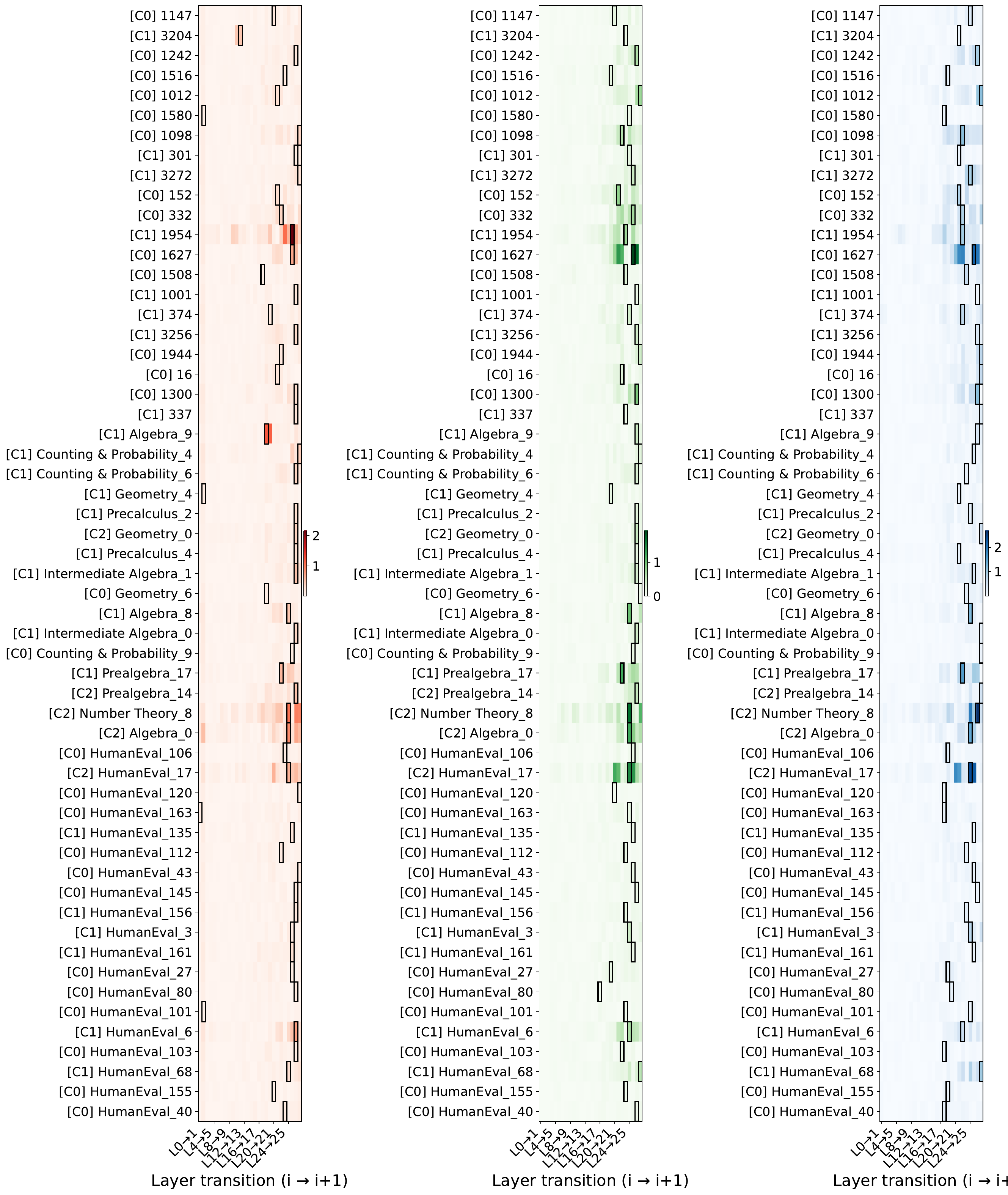}
    \caption{Layer-wise $|\Delta|$ heatmaps for SB, BOS, and OC across all traces (rows). Black boxes mark the peak transition layer for each trace. Darker colors indicate larger magnitude of change.}
    \label{fig:delta_heatmap}
\end{figure}

The second robust aggregate finding concerns the location of \emph{critical transition layers}. We compute first-order differences ($\Delta$) of each decomposition component across adjacent layers and locate the \emph{peak transition layer} for each trace, defined as the layer transition at which the component changes most in absolute value (\autoref{fig:delta_heatmap}).

Despite the heterogeneity observed in case-level clustering, the peak transition layers are remarkably consistent: median peaks concentrate in layers $20$--$26$ across all clusters (\autoref{tab:peak_layers}). For BOS and OC in particular, the peak transitions cluster tightly around layers 23--26 (std $\leq 4.0$), indicating that the model's final ``decision moment,'' i.e., the layer at which the context-versus-bias tradeoff is resolved, is largely invariant to the specific failure case or profile shape.

This consistency has a direct practical implication: targeted interventions for correcting failure-inducing biases (e.g., activation editing or attention head pruning) should focus on the \textbf{final 6--8 layers} of the network, where the decisive attribution transitions reliably occur.

\begin{table}[htbp]
\centering
\caption{Median peak transition layer per cluster (layer $i \to i+1$).}
\label{tab:peak_layers}
\small
\begin{tabular}{lccc}
\toprule
\textbf{Cluster} & \textbf{$\Delta$SB peak} & \textbf{$\Delta$BOS peak} & \textbf{$\Delta$OC peak} \\
\midrule
Cluster 0 ($n{=}13$) & $26.0 \pm 6.8$ & $26.0 \pm 2.2$ & $24.0 \pm 2.1$ \\
Cluster 1 ($n{=}3$)  & $24.0 \pm 0.0$ & $24.0 \pm 0.0$ & $24.0 \pm 0.9$ \\
Cluster 2 ($n{=}7$)  & $20.0 \pm 10.0$ & $24.0 \pm 2.1$ & $23.0 \pm 4.0$ \\
\bottomrule
\end{tabular}
\end{table}

\subsection{Cross-Model Comparison of Attribution Decomposition}

\begin{figure}[h]
    \centering
    \includegraphics[width=\columnwidth]{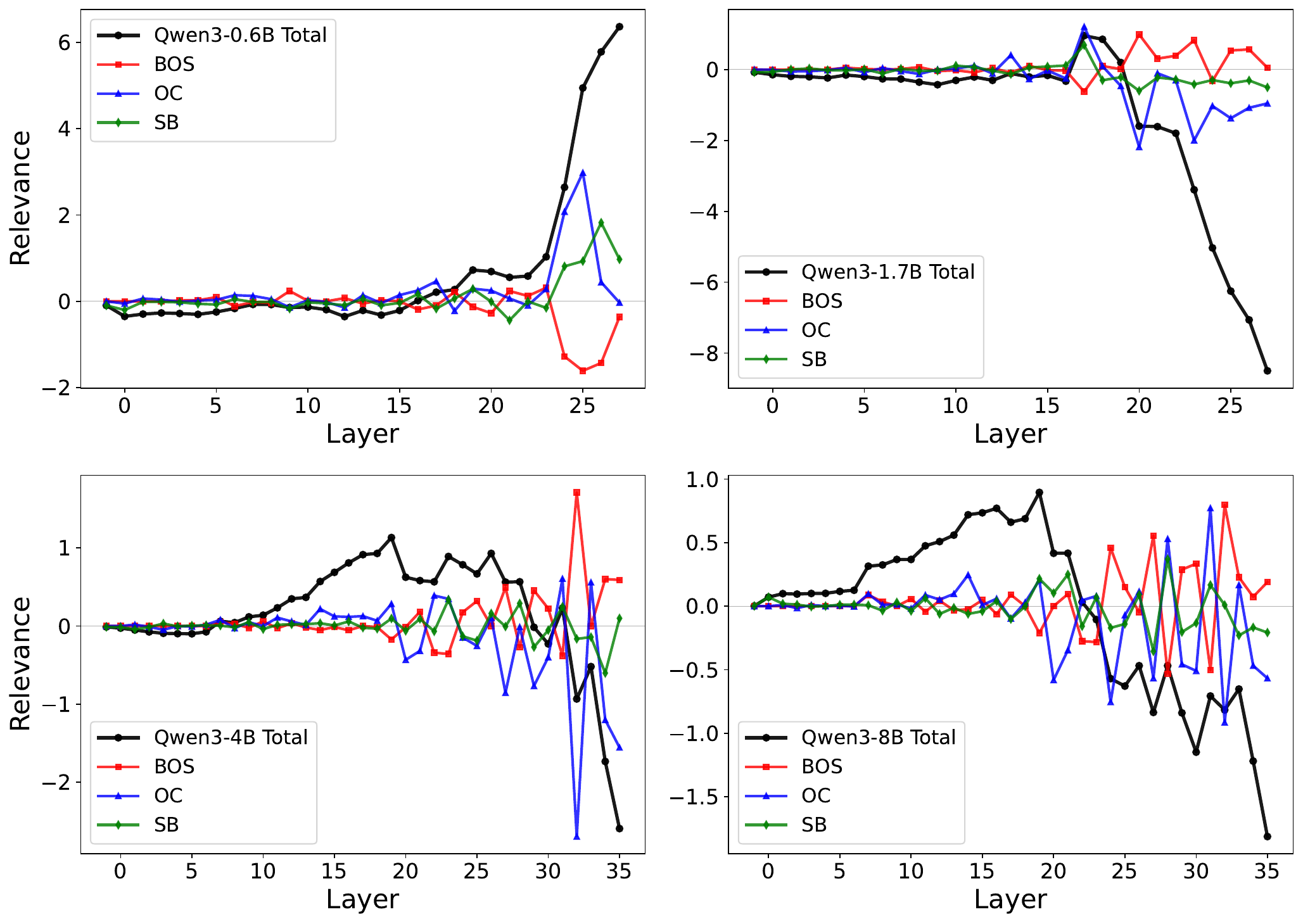}
    \caption{Cross-model attribution decomposition comparison for a representative failure case. Each panel shows the layer-wise decomposition (Total, SB, BOS, OC) for a different Qwen3 model size.}
    \label{fig:cross_model_decomp}
\end{figure}

Using attribution graphs from multiple Qwen3 model sizes (0.6B, 1.7B, 4B, 8B) computed on the same 23 NC-IA failure cases, we compare how the decomposition structure changes with model scale. For each failure case, we extract the same three components (SB, BOS, OC) and plot their layer-wise trajectories for all four model sizes.

Representative cross-model decomposition comparisons are shown in \autoref{fig:cross_model_decomp}. Taking the MATH case \emph{Number Theory\_8} as an illustrative example, the layer-averaged decomposition values across model sizes are:

\begin{table}[htbp]
\centering
\caption{Cross-model decomposition for a representative MATH failure case (\emph{Number Theory\_8}). Values are layer-averaged.}
\label{tab:cross_model_example}
\small
\begin{tabular}{lcccc}
\toprule
\textbf{Component} & \textbf{0.6B} & \textbf{1.7B} & \textbf{4B} & \textbf{8B} \\
\midrule
Total relevance & $0.693$ & $-1.288$ & $0.168$ & $-0.036$ \\
BOS contribution & $-0.145$ & $0.100$ & $0.086$ & $0.041$ \\
Other contribution & $0.246$ & $-0.294$ & $-0.147$ & $-0.074$ \\
Self bias & $0.118$ & $-0.097$ & $-0.008$ & $-0.016$ \\
\bottomrule
\end{tabular}
\end{table}

Several consistent patterns emerge across failure cases:
\begin{itemize}[leftmargin=*]
    \item \textbf{Total relevance shifts toward negative values with scale.} For corrected cases, the total relevance shifts from positive (favoring the incorrect token) in the 0.6B model to negative (favoring the correct alternative) in larger models, reflecting the corrected logit difference. For the above example, the 0.6B model has a positive total relevance of $0.693$, whereas the 1.7B model already reverses to $-1.288$.
    \item \textbf{BOS contribution diminishes with scale.} The absolute BOS contribution generally decreases from 0.6B to 8B ($0.145 \to 0.041$ in the above example), suggesting that larger models are less reliant on the attention-sink shortcut.
    \item \textbf{Context integration sign flips for corrected cases.} When a larger model corrects the failure, the OC component typically flips sign (from positive to negative), indicating that the larger model assigns contextual relevance that now supports the correct token rather than the incorrect one.
\end{itemize}

These cross-model findings complement the input-level attribution scaling analysis in~\autoref{fig:weak_strong_rel_breakdown} (main text) by revealing that the improvement in larger models is not merely a surface-level token reweighting, but reflects deeper structural changes in how relevance is composed and propagated across the transformer stack.

\subsection{Summary}

This analysis reveals a tension between case-level and aggregate-level structure in attribution graphs. At the case level, the data are inherently multi-dimensional: relevance profile shape and source composition are orthogonal (ARI $\approx 0$), and no single clustering scheme captures the structural diversity of failure cases. This underscores the challenge of building automated failure taxonomies from attribution graphs at the current scale and with current decomposition granularity. At the aggregate level, however, two robust patterns emerge that are consistent across all traces: (1)~a three-phase functional specialization (early bias $\to$ mid-layer context integration $\to$ late-layer bias amplification), and (2)~critical attribution transitions concentrated in the final 6--8 layers. These aggregate regularities provide actionable guidance for model interventions and connect mechanistically to the behavioral failure patterns identified in the main text.

\end{document}